\documentclass{bmvc2k}

\usepackage{booktabs}
\usepackage{multirow}
\usepackage[table]{xcolor}
\usepackage{booktabs}
\usepackage{makecell}
\usepackage[table]{xcolor}
\usepackage{makecell}
\usepackage{array}
\usepackage{amssymb}

\usepackage{graphicx}
\usepackage{booktabs}

\usepackage{siunitx}
\title{Inverse-and-Edit: Simple and Effective Framework for Fast Image Editing}

\addauthor{Ilia Beletskii}{ibeletskiy@hse.ru}{1, 2}
\addauthor{Andrey Kuznetsov}{kuznetsov@fusionbrainlab.com}{2, 3}
\addauthor{Aibek Alanov}{alanov.aibek@gmail.com}{1, 2, 3}

\addinstitution{
 HSE University, \\
 Moscow, Russia
}
\addinstitution{
 FusionBrain Lab, AXXX, \\
 Moscow, Russia
}
\addinstitution{
 Artificial Intelligence Institute, Innopolis University\\
 Innopolis, Russia
}

\runninghead{Beletskii, Kuznetsov, Alanov}{Inverse-and-Edit}

\begin{document}

\maketitle

\begin{figure}[h]
\centering
\includegraphics[width=\linewidth]{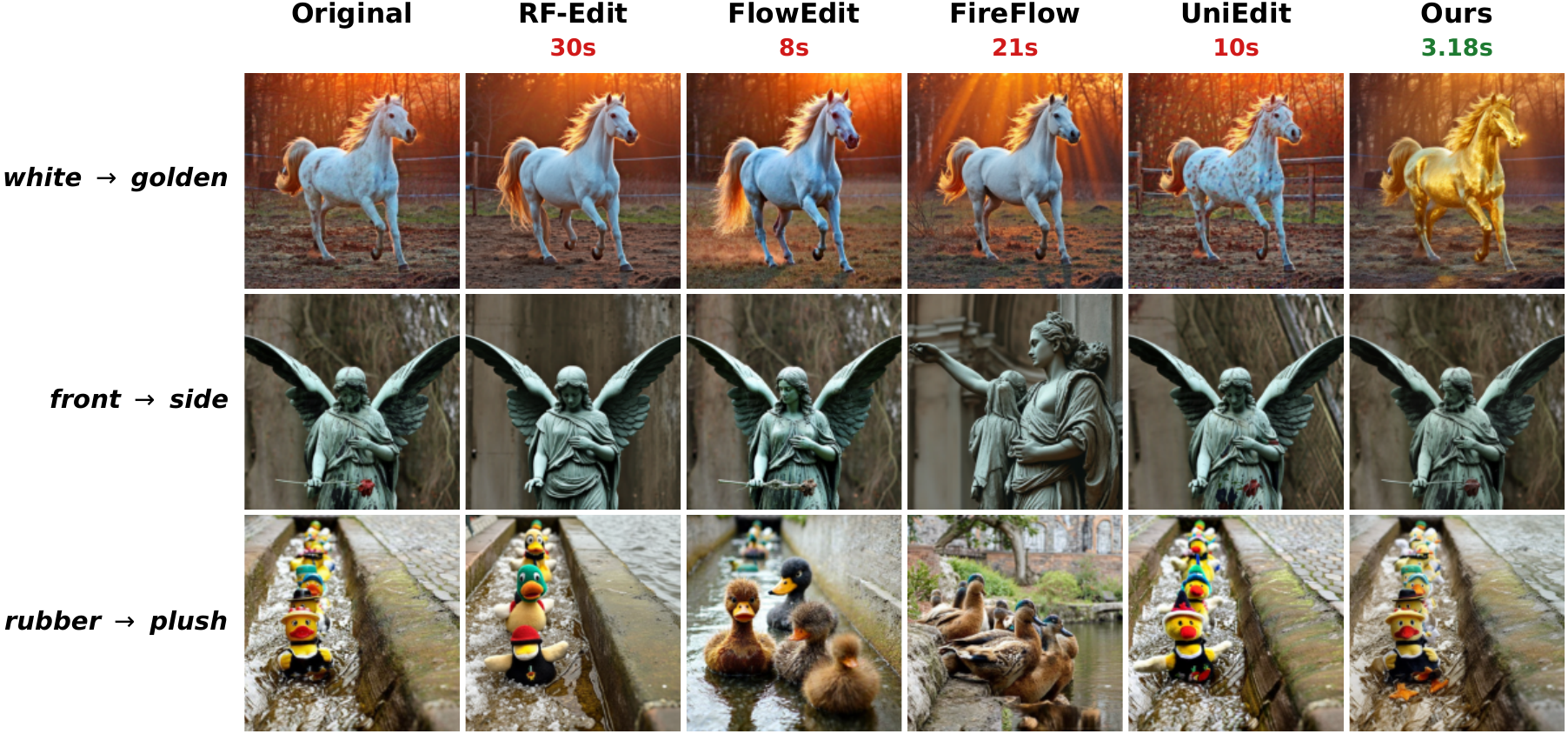}
    \caption{\textbf{Qualitative comparison on Z-Image-Turbo (flow-based)} against
RF-Edit, FlowEdit, FireFlow, and UniEdit-Flow (inference time below each column;
one edit per row). Our method applies each edit while preserving structure, and is
the fastest editor at 3.18\,s.}
    \label{fig:image_edits_examples_flow}
\end{figure}
\begin{abstract}
Diffusion-based text-to-image models have recently been distilled into few-step samplers and fast generators, reducing inference cost while maintaining high image quality. However, text-guided editing with such accelerated backbones remains challenging: inversion in a small number of steps is unreliable, local errors accumulate along the trajectory, and existing fast-editing methods typically rely on backbone-specific inversion procedures, which limits their transferability. In this work, we propose a simple and universal framework for text-guided image editing with arbitrary accelerated diffusion and flow-matching models. Instead of optimizing local, per-timestep objectives, we directly minimize a global editing objective over the entire inversion-generation trajectory. The framework fine-tunes only 1.5-7 \% of the base-model parameters through two lightweight LoRA adapters (forward and backward), trained on fewer than 10 K triplets from the publicly available HQ-Edit dataset. We validate our method on three architecturally distinct backbones: consistency-distilled SD1.5 LCM-Dreamshaper, adversarial-distilled SDXL-Turbo, and rectified-flow-distilled Z-Image-Turbo. On PIE-Bench, our method achieves the strongest editing alignment within each backbone family while ensuring a strong level of content preservation. Inference adds no extra steps beyond the standard inversion and editing passes. The project page with visual results is available at iliabeletskii.github.io/inverse-and-edit. 
\end{abstract}

\section{Introduction}
\label{sec:intro}
Diffusion-based text-to-image models~\citep{ho2020denoisingdiffusionprobabilisticmodels,song2021scorebasedgenerativemodelingstochastic} have become a standard tool for high-fidelity image synthesis. Beyond generation, a central downstream capability is text-guided image editing: given an input image and a target prompt, the model should apply prompt-consistent modifications while keeping unrelated content (identity, layout, background) unchanged. Achieving this balance—strong editability with faithful preservation—typically requires carefully controlling the sampling trajectory.

Editing with diffusion pipelines achieves high image quality at the cost of significant inference time. To address this latency, recent work explores distilled diffusion models. The backbones of such methods can be broadly categorized into two groups: consistency-based~\citep{song2023consistencymodels, ren2024hypersdtrajectorysegmentedconsistency} and fast generator-based approaches~\citep{sauer2024fasthighresolutionimagesynthesis, yin2024onestepdiffusiondistributionmatching}. Consistency-based methods preserve a solver-based view of diffusion dynamics by optimizing few-step solvers for the underlying differential equation (e.g., Latent Consistency Models ~\citep{luo2023latentconsistencymodelssynthesizing}, Hyper-SD ~\citep{ren2024hypersdtrajectorysegmentedconsistency}, Invertible Consistency Distillation ~\citep{starodubcev2024invertibleconsistencydistillationtextguided}), whereas generator-based models (e.g., SDXL-Turbo~\citep{sauer2023adversarialdiffusiondistillation}) train a neural network to map standard Gaussian noise \(z\) directly to high-quality images in a few steps.

Despite the importance of editing speed, most methods were originally formulated for full-step models, where editing proceeds in two stages: inversion recovers a terminal latent $x_T$ that reconstructs the input $x_0$ under $y_{\text{src}}$, and editing re-denoises that latent under $y_{\text{trg}}$. This works because adjacent states are local ($x_t \approx x_{t-1}$), making DDIM-style inversion accurate, so full-step editors focus on the generation stage, steering it via attention or guidance (e.g., Prompt-to-Prompt~\cite{hertz2022prompttopromptimageeditingcross}, MasaCTRL~\cite{cao2023masactrltuningfreemutualselfattention}, Guide-and-Rescale~\cite{titov2024guideandrescaleselfguidancemechanismeffective}).
As distillation progressed, several works adapted full-step editing to accelerated backbones (e.g., InfEdit~\cite{xu2023inversionfreeimageeditingnatural}, TurboEdit~\cite{deutch2024turboedittextbasedimageediting}, iCD~\cite{starodubcev2024invertibleconsistencydistillationtextguided}, GNRi~\cite{samuel2025lightningfastimageinversionediting}). However, inversion becomes the bottleneck: few-step models traverse the noise range in large jumps, so $x_t\approx x_{t-1}$ breaks down and small inversion errors compound into poor reconstruction or drifting edits.
Existing fast editors handle this in backbone-specific ways, such as replacing inversion, adding per-image latent optimization, or jointly training a dedicated inversion model (Sec.~\ref{sec:rel_works}). These either erode the speedup or couple tightly to one distillation recipe, making transfer non-trivial.

A core limitation of prior accelerated editors is that they typically optimize ``local'' surrogates (e.g., enforcing step-wise consistency) and hope this transfers to the \emph{global} objective of the full pipeline: accurate reconstruction under the source prompt and faithful editing under the target prompt. In few-step backbones, however, the trajectory consists of large jumps, so small mismatches quickly compound and the final edited image can drift even if each step is locally reasonable. Our key motivation is that accelerated models make it feasible to \textbf{differentiate through the entire inversion--generation procedure} and optimize what we ultimately care about.
We therefore treat accelerated editing as a \textbf{single end-to-end computation graph}. Given an image $x_0$, a source prompt $y_{\text{src}}$, and a target prompt $y_{\text{trg}}$, we (i) invert $x_0$ to a terminal latent $x_T$ under $y_{\text{src}}$, and (ii) denoise $x_T$ under $y_{\text{trg}}$ to obtain the edited result $\hat{x}_0$. Instead of enforcing local consistency between neighboring states, we fine-tune two lightweight LoRA adapters---\textbf{forward} (for inversion/noising) and \textbf{backward} (for generation/denoising)---by backpropagating through the full inversion--generation pipeline.
Our training objective explicitly encodes three competing requirements of fast editing: \textbf{(1) reconstruction/content preservation} via a perceptual reconstruction loss, \textbf{(2) edit strength and prompt faithfulness} via a directional CLIP objective between $(x_0,\hat{x}_0)$ and $(y_{\text{src}},y_{\text{trg}})$, and \textbf{(3) well-posed inversion} via a prior-matching regularizer that prevents the inverted latent from becoming under-noised. Crucially, the short trajectory of few-step backbones makes this end-to-end optimization practical, and the resulting editor remains \textbf{backbone-agnostic}: the same recipe applies across distilled diffusion samplers, fast generators, and flow-based models without adding extra inference steps at test time.

In this work, we make three main contributions:
\begin{enumerate}
    \item \textbf{Joint inversion--generation optimization for accelerated
editing.} We fine-tune two lightweight LoRA adapters (forward for
inversion, backward for generation) by backpropagating through the
full inversion--generation cycle of a frozen few-step backbone, with a
terminal-prior regularizer that closes the under-noising shortcut of
this joint objective.
    \item \textbf{Backbone- and distillation-agnostic recipe across U-Net and DiT.}
    The same recipe applies unchanged across three distinct distillation paradigms
    (consistency-distilled SD1.5, adversarially distilled SDXL-Turbo, and
    distribution-matching-distilled Z-Image-Turbo) and two architecture families
    (U-Net and a single-stream DiT), with no backbone-specific inversion machinery and
    no extra inference steps at test time.
    \item \textbf{Strong quality--speed trade-off.} Across all three
backbones our method achieves the strongest edit alignment within each
backbone family at comparable or lower inference cost, and preserves
more than prior editors at matched edit strength. On the flow
backbone, a human preference study against competing flow-based
editors corroborates these gains in preservation and overall
preference.
\end{enumerate}
\section{Related work}
\label{sec:rel_works}
\paragraph{Editing with full-step diffusion models.} 
Most full-step methods begin with inversion, i.e., approximating a latent state of the source image at a chosen timestep using a pretrained model. This state then initializes a new, target prompt-conditioned sampling trajectory. Optimization-based approaches~\citep{miyake2024negativepromptinversionfastimage, Mokady_2023_CVPR, zuo2025precise, Wu_2026} perform per-sample inversion optimization, which involves additional, computationally expensive iterations during the editing process. These methods improve inversion quality by optimizing prompt embeddings. 
Attention-based approaches~\citep{cao2023masactrltuningfreemutualselfattention,hertz2022prompttopromptimageeditingcross} demonstrate strong performance but may lack fine-grained controllability.
\paragraph{Editing with accelerated diffusion models.}
Distilled backbones reduce sampling cost but may degrade trajectory fidelity, making
standard inversion less reliable, since deterministic inversion is not guaranteed under
stochastic sampling or non-invertible updates. Prior accelerated editors differ chiefly
in whether, and how, they recover an editable starting point. Some strengthen inversion
on the distilled backbone: on SDXL-Turbo, GNRi~\citep{samuel2025lightningfastimageinversionediting}
and ReNoise~\citep{garibi2024renoiserealimageinversion} refine the inverted latent through
per-image optimization. Others reduce reliance on explicit inversion:
TurboEdit~\citep{deutch2024turboedittextbasedimageediting} alters the sampling dynamics via
time-shifting and Delta Denoising Score~\citep{hertz2023deltadenoisingscore} updates, while
PostEdit~\citep{tian2025posteditposteriorsamplingefficient} applies energy-based guidance
during editing, and InstantEdit~\cite{instantedit} pairs a
backbone-specific rectified-flow inversion with inference-time
latent injection. A third group avoids inversion altogether:
InfEdit~\citep{xu2023inversionfreeimageeditingnatural}, on a DreamShaper-LCM backbone, adopts
a virtual inversion that starts from generic Gaussian noise and combines
MasaCTRL~\citep{cao2023masactrltuningfreemutualselfattention} with
Prompt-to-Prompt~\citep{hertz2022prompttopromptimageeditingcross}, and
ChordEdit~\cite{lu2026chordedit} recasts one-step editing (SD-Turbo, SwiftBrushv2) as optimal
transport between source- and target-prompt distributions via a \emph{Chord Control Field}
traversed in a single step. In contrast to fixed transport rules or attention edits, our method learns lightweight
forward and backward adapters end-to-end on a frozen backbone, obtaining the editing
direction from the jointly trained inversion and generation passes. Because it relies only
on properties common to accelerated diffusion and flow samplers, namely a differentiable
few-step sampling loop and standard adapter-injectable layers, rather than on any
backbone-specific inversion or attention mechanism, the same recipe applies unchanged
across U-Net and DiT backbones.

\paragraph{Fine-tuning-based fast editors.}
A separate line of work trains dedicated weights up front so that editing is fast at
inference, rather than relying on per-image optimization or attention control alone.

iCD~\cite{starodubcev2024invertibleconsistencydistillationtextguided} trains a forward
(inversion) and a reverse (generation) consistency model under a multi-boundary
objective, obtaining its inversion ability through consistency distillation on the
diffusion probability-flow ODE and, for editing, relying on Prompt-to-Prompt attention
control. It is demonstrated only on consistency-distilled diffusion backbones
(SD1.5, SDXL), leaving open whether its inversion extends to adversarially distilled or
flow-matching models. Our method needs neither: we attach lightweight LoRA adapters to
a frozen backbone and treat it as a black box, so the same recipe extends to the
adversarial and flow-matching models iCD leaves untested.

SwiftEdit~\cite{Nguyen_2025_CVPR} takes a heavier route: on top of the one-step
SwiftBrushv2 generator~\cite{dao2024swiftbrushv2} (an SD2.1-distilled U-Net,
$\sim\!866$M parameters), it fully trains a second U-Net of the same scale as a
dedicated one-step inversion network in two stages, with an IP-Adapter-style
conditioning branch and an attention-rescaling editing mechanism, yielding instant
(${\sim}0.2$\,s) edits. The framework can be re-instantiated on other one-step
generators (e.g., DMD2, SwiftBrush), but all remain U-Net-based (SD1.5 or SD2.1) and
each re-pays the full two-stage training; no SDXL or flow/DiT backbone is demonstrated. Wu et al.~\cite{wu2024turboeditinstanttextbasedimage} similarly train a separate
reconstruction-only inversion encoder on SDXL-Turbo, editing at
inference via caption manipulation. Ours is far lighter: instead of fully training a second generator-scale network in two
stages, we adapt two lightweight LoRA adapters---a forward (inversion) and a backward
(generation) adapter---on a single frozen backbone, jointly in one stage for just
$300$--$600$ iterations on fewer than $10$K triplets ($1.5$--$7\%$ of the base-model
parameters), with no IP-Adapter or attention control at inference (A component-by-component comparison to SwiftEdit~\cite{Nguyen_2025_CVPR} is given in Appendix~\ref{app:swiftedit}).
The result is a single recipe that transfers unchanged across U-Net and
transformer-based (DiT) backbones, including the flow-matching models neither iCD nor
SwiftEdit demonstrates. Our scheme relates to reward fine-tuning
(ReFL~\cite{xu2023imagerewardlearningevaluatinghuman}, DRaFT~\cite{clark2024directlyfinetuningdiffusionmodels},
ShortFT~\cite{guo2025shortftdiffusionmodelalignment}), which backpropagates through the generation
pass alone; we optimize the joint inversion--generation cycle, whose
under-noising shortcut has no counterpart there. Concurrent
BiFM~\cite{dai2026bifmbidirectionalflowmatching} learns both directions but trains from scratch,
whereas we adapt a frozen pretrained backbone.
\paragraph{Editing with flow-based models.}
Recent text-to-image systems increasingly adopt flow-matching / rectified-flow
objectives~\cite{lipman2022flow, liu2022flow}, parameterizing a learned velocity field
as an ODE (e.g., SD3~\cite{esser2024scaling} and FLUX). Editing again hinges on
inversion, which is difficult even in the many-step regime, since the learned velocity
field makes naive ODE-reversal imprecise. Recent flow editors improve this in different
ways: RF-Edit~\cite{wang2025tamingrectifiedflowinversion} uses a second-order solver and
shares self-attention features across the inversion and editing passes;
FireFlow~\cite{deng2025fireflow} attains second-order accuracy at first-order cost in
eight steps; UniEdit-Flow~\cite{jiao2025unieditflowunleashinginversionediting} pairs a
predictor--corrector inversion with a region-aware sampler; and
FlowEdit~\cite{kulikov2025floweditinversionfreetextbasedediting} avoids inversion
altogether but still needs many steps ($28$--$50$ on FLUX/SD3). All operate with tens of
steps rather than the few-step regime of aggressively distilled backbones such as
Z-Image-Turbo, and several add inference-time machinery (feature-sharing in RF-Edit,
region-aware sampling in UniEdit-Flow). Operating directly on a distilled
rectified-flow backbone (Z-Image-Turbo), our method achieves the fastest inference among
flow-based editing methods, requiring only a few sampling steps rather than the tens used
by prior flow editors.
\paragraph{Instruct-based editing.}
A growing line of work replaces source/target prompt pairs with editing instructions,
learning a direct image$+$instruction$\rightarrow$edit mapping rather than inverting and
regenerating (e.g., InstructPix2Pix~\citep{brooks2023instructpix2pixlearningfollowimage},
FLUX Kontext~\citep{labs2025flux1kontextflowmatching}). These typically require heavy
fine-tuning of a large text-to-image model and large supervised datasets of edited
ground-truth images. Our method needs neither: we keep a frozen backbone and train
lightweight adapters on fewer than $10$K prompt-pair triplets, with no edited
ground-truth targets.
\section{Preliminaries}
\label{sec:preliminaries}


\paragraph{Diffusion and flow models.}
We build on a latent text-to-image model that operates in the VAE latent space. Given an
image $x_0$, we encode it into $z_0 = \mathrm{enc}(x_0)$. In the diffusion case, the
forward process produces noisy latents $\{z_t\}_{t=1}^{T}$ via a variance schedule
$\{\alpha_t\}$ with $\bar\alpha_t = \prod_{s=1}^{t}\alpha_s$: $z_t = \sqrt{\bar\alpha_t}\,z_0 + \sqrt{1-\bar\alpha_t}\,\varepsilon,$ $\varepsilon \sim \mathcal{N}(0,I).$
A denoiser $\varepsilon_\theta(z_t,t,y)$ (conditioned on text $y$ through cross-attention)
is trained by noise prediction,
\begin{equation}
  \label{eq:diffusion_loss}
  \mathcal{L}(\theta) = \mathbb{E}_{t,z_0,\varepsilon}
  \big[\,\|\varepsilon_\theta(z_t,t,y)-\varepsilon\|_2^2\,\big].
\end{equation}
At inference, generation starts from $z_T \sim \mathcal{N}(0,I)$ and applies a
deterministic DDIM-style update (equivalently, an ODE discretization) to map
$z_t \mapsto z_{t-1}$:
\begin{equation}
  z_{t-1} = \sqrt{\tfrac{\bar\alpha_{t-1}}{\bar\alpha_t}}\,z_t
  + \Big(\sqrt{1-\bar\alpha_{t-1}}
  - \sqrt{\tfrac{\bar\alpha_{t-1}}{\bar\alpha_t}}\sqrt{1-\bar\alpha_t}\Big)
  \varepsilon_\theta(z_t,t,y).
  \label{eq:ddim}
\end{equation}
Rectified-flow models replace the noise-prediction view with a learned velocity field. A
network $v_\theta(z_t,t,y)$ is trained to match the velocity of a straight path between
data and noise, $z_t = (1-t)\,z_0 + t\,\varepsilon$, by regressing the target
$\varepsilon - z_0$; sampling then integrates the probability-flow ODE
$\mathrm{d}z_t = v_\theta(z_t,t,y)\,\mathrm{d}t$ from $z_T$ to $z_0$. A single
deterministic Euler step over the interval $t \to t'$ gives
\begin{equation}
  z_{t'} = z_t + (t'-t)\,v_\theta(z_t,t,y).
  \label{eq:flow}
\end{equation}
Although their training objectives differ, both~\eqref{eq:ddim} and~\eqref{eq:flow}
define a single deterministic step between adjacent noise levels along the sampling
trajectory. Our method operates only on this step, which is what allows the same forward
and backward adapters to apply to diffusion and flow backbones alike.
\paragraph{Distilled diffusion and flow models.}
Distillation cuts sampling from tens of steps to a few,
$0=t_0<t_1<\dots<t_M=T$ with $M$ small. Our experiments span three paradigms, each
replacing the noise-prediction loss~\eqref{eq:diffusion_loss} with a few-step objective:
\begin{align}
  \text{CD:}\;\;
  &\mathcal{L}_{\mathrm{CD}}=\big\|f_\theta(z_{t_m},t_m,y)-f_{\theta^-}(z_{t_{m-1}},t_{m-1},y)\big\|,
  \label{eq:cd}\\
  \text{ADD:}\;\;
  &\mathcal{L}_{\mathrm{ADD}}=\mathcal{L}_{\mathrm{adv}}\!\big(D_W(\hat{x}_0)\big)+\lambda\,\mathcal{L}_{\mathrm{dist}}(\hat{x}_0,x_0^{\mathrm{tch}}),
  \label{eq:add}\\
  \text{DMD:}\;\;
  &\nabla_\theta\mathcal{L}_{\mathrm{DMD}}\propto\mathbb{E}\big[\,s_{\mathrm{fake}}-s_{\mathrm{real}}\,\big],
  \label{eq:dmd}
\end{align}
where $f_\theta$ is a consistency function (with EMA target copy $f_{\theta^-}$) mapping a
PF-ODE point to its endpoint; $\hat{x}_0$ is the student's few-step output, $D_W$ a
discriminator, $\mathcal{L}_{\mathrm{dist}}$ a distillation term toward the teacher output
$x_0^{\mathrm{tch}}$ with weight $\lambda$; and
$s_{\mathrm{real}}\!=\!\nabla\log p_{\mathrm{real}}$,
$s_{\mathrm{fake}}\!=\!\nabla\log p_{\mathrm{fake}}$ are the teacher and student scores, so
the gradient drives $p_{\mathrm{fake}}\!\to\!p_{\mathrm{real}}$. These instantiate,
respectively, consistency distillation (LCM~\citep{luo2023latentconsistencymodelssynthesizing},
the DreamShaper-LCM we use on SD1.5),
adversarial diffusion distillation (SDXL-Turbo~\citep{sauer2023adversarialdiffusiondistillation},
our SDXL backbone), and distribution-matching distillation
(\citep{yin2024onestepdiffusiondistributionmatching}; Z-Image-Turbo, our flow backbone).
Whatever the objective, collapsing many steps into a few makes each transition span a large
interval $\Delta_m=t_m-t_{m-1}$, so adjacent latents cease to be local,
\begin{equation}
  z_{t_{m-1}}\not\approx z_{t_m},\qquad
  \varepsilon_\theta(z_{t_m},t_m,y)\not\approx\varepsilon_\theta(z_{t_{m-1}},t_m,y),
  \label{eq:largejump}
\end{equation}
and the near-locality that inversion-based editing relies on no longer holds.
\paragraph{Editing as inversion and generation.}
Both deterministic steps~\eqref{eq:ddim} and~\eqref{eq:flow} map a latent at one noise
level to the adjacent one. Iterating the inversion direction ($t_{m-1}\!\to\!t_m$) from
$z_{t_0}=\mathrm{enc}(x_0)$ defines the terminal latent, and iterating the generation
direction ($t_m\!\to\!t_{m-1}$) from it reconstructs the source:
\begin{equation}
  z_{t_m}\!=\!\mathrm{step}(z_{t_{m-1}};t_{m-1}\!\to\!t_m,y),\;\; m{=}1{:}M,
  \quad
  \hat{z}_{t_{m-1}}\!=\!\mathrm{step}(\hat{z}_{t_m};t_m\!\to\!t_{m-1},y),\;\; m{=}M{:}1,
  \label{eq:invgen-rec}
\end{equation}
where $\mathrm{step}(\cdot)$ is the backbone update~\eqref{eq:ddim}/\eqref{eq:flow},
$z_T\triangleq z_{t_M}$, and $\hat{z}_{t_M}\triangleq z_T$. We adopt the standard two-stage
pipeline: \textbf{inversion} runs the left recurrence under the \textbf{source prompt}
$y_{\mathrm{src}}$ to obtain $z_T$, and \textbf{editing generation} runs the right
recurrence from the same $z_T$ under either prompt,
\begin{equation}
  z_T=\mathrm{Inv}(x_0,y_{\mathrm{src}}),\quad
  \mathrm{dec}\!\big(\mathrm{Gen}(z_T,y_{\mathrm{src}})\big)\approx x_0,\quad
  \hat{x}_{\mathrm{trg}}=\mathrm{dec}\!\big(\mathrm{Gen}(z_T,y_{\mathrm{trg}})\big).
  \label{eq:recon-edit}
\end{equation}
Standard DDIM inversion solves~\eqref{eq:recon-edit} accurately for full-step samplers but
degrades in the few-step regime~\eqref{eq:largejump}; we therefore learn $\mathrm{Inv}$ and
$\mathrm{Gen}$ jointly (as adapters $f_\phi,g_\theta$ in Sec.~\ref{sec:method}) rather than
approximate them, so $z_T$ is the latent the generation stage is calibrated to denoise.
\section{Method}
\label{sec:method}



For accelerated diffusion backbones, inversion is often the dominant failure mode: few-step transitions amplify small inversion errors, which then propagate into the edited result. Prior fast editing methods frequently address this by introducing backbone-specific inversion tricks or per-image optimization, making transfer across distilled models non-trivial. Our key observation is that the small number of steps in accelerated samplers makes \textbf{end-to-end training over the entire inversion--generation loop feasible}. We therefore directly fine-tune a lightweight adaptation for editing so that (i) inversion produces a terminal latent that behaves like valid model noise, and (ii) denoising from that latent reliably reconstructs the source under $y_\text{src}$ and performs the intended change under $y_\text{trg}$. This framing makes the reconstruction--edit trade-off explicit: we simultaneously enforce high-fidelity content preservation and strong, prompt-consistent edits.


We adapt a pre-trained distilled backbone $D_W$ for editing using two lightweight LoRA parameter sets. Concretely, we instantiate \textbf{two copies of the same frozen network} with different LoRA weights: a \emph{forward} model $F_\phi(\cdot)=D_{W+\Delta W_\phi}(\cdot), \text{ used only for inversion under } y_\text{src},$
and a \emph{backward} model $G_\theta(\cdot)=D_{W+\Delta W_\theta}(\cdot), \text{ used for denoising under } y_\text{src}\text{ or }y_\text{trg}.$
The role of $F_\phi$ is to map $x_0$ to a terminal latent $z_T$ that (a) preserves enough information to reconstruct the source, but (b) remains statistically compatible with the backbone’s noise prior, so that few-step denoising remains prompt-sensitive. The role of $G_\theta$ is to make the denoising dynamics robust to the (learned) inversion distribution produced by $F_\phi$, improving stability and edit fidelity. Training both adapters jointly aligns inversion and generation without modifying the backbone weights.


\subsection{Latent representation alignment}
Optimizing only the inversion mapping $F_\phi$ is possible but tends to be unstable: the model can exploit shortcut solutions that reconstruct well yet yield a terminal latent distribution that the (fixed) denoiser is not calibrated for, degrading editing. We therefore optimize $F_\phi$ and $G_\theta$ \textbf{jointly}, allowing the backward model to adapt to the learned inversion distribution while still keeping the shared backbone weights $W$ frozen. This ``two-sided'' alignment makes training more sample-efficient and improves robustness across prompts and images, because inversion and generation are calibrated to each other under the same few-step schedule.

\subsection{General framework for image editing}
\begin{figure}
\centering
\includegraphics[width=\linewidth]{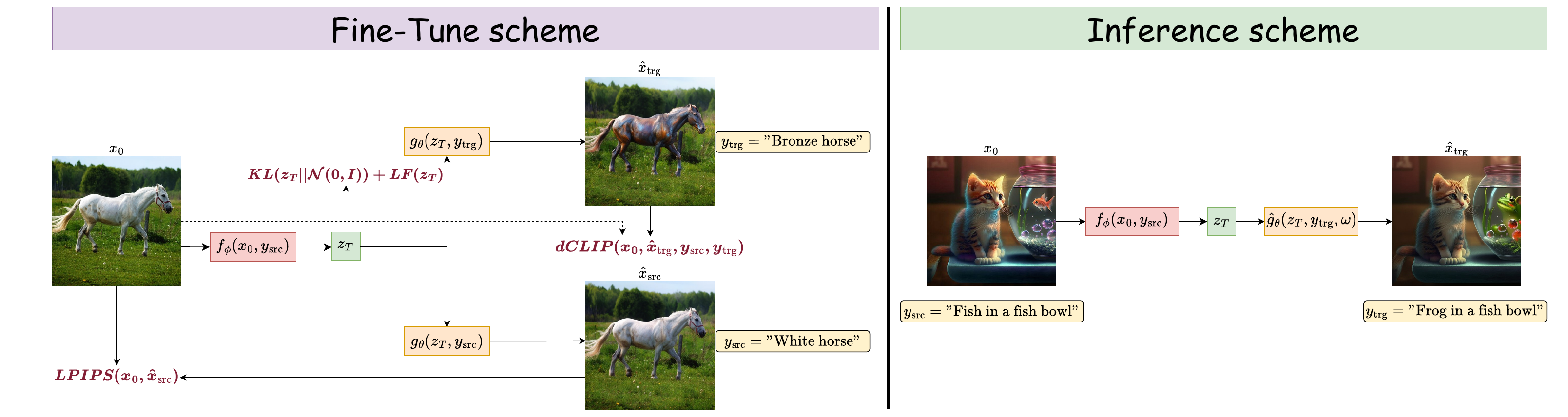}
    \caption{\textbf{Overview of our framework.}
\emph{Left (fine-tuning):} the forward adapter $f_\phi$ inverts $x_0$ to a
terminal latent $z_T$; the backward adapter $g_\theta$ denoises $z_T$ under
$y_{\mathrm{src}}$ (reconstruction, scored by LPIPS) and under $y_{\mathrm{trg}}$
(edit, scored by directional CLIP). Both adapters are trained jointly end-to-end
with a prior regularizer $\mathrm{KL}+\mathrm{LF}$ on $z_T$ and the backbone
frozen (Eq.~\ref{eq:final_objective}). \emph{Right (inference):} the same $f_\phi$
inverts, then guided $\hat{g}_\theta(z_T,y_{\mathrm{trg}},\omega)$ produces the
edit with no auxiliary machinery (Eq.~\ref{eq:editing_inference}).}
    \label{fig:framework}
\end{figure}

We operate in the VAE latent space: for an input image $x_0$ we compute $z_0=\mathrm{enc}(x_0)$. Our editing pipeline consists of a forward (inversion/noising) pass that maps $z_0$ to a terminal latent $z_T$, followed by a backward (generation/denoising) pass that maps $z_T$ back to an image under a chosen prompt (Fig. ~\ref{fig:framework}). Since we target accelerated backbones with a small number of inference steps, we explicitly discretize the trajectory into $M$ noise levels $0=t_0<t_1<\dots<t_M=T$ and train through the entire forward--backward loop end-to-end.

\paragraph{Reconstruction.}
We formulate reconstruction as a closed-loop process executed with a small number of steps. We discretize the diffusion trajectory into $M$ noise levels $0=t_0<t_1<\dots<t_M=T$ consistent with the accelerated backbone. Given an input image $x_0$ and its source prompt $y_\text{src}$, we first encode $x_0$ into $z_{t_0}=z_0=\mathrm{enc}(x_0)$ and apply the forward (inversion/noising) model $F_\phi$ step-by-step:
\begin{equation}
\label{eq:forward_step}
z_{t_m}=F_\phi\!\big(z_{t_{m-1}},\,t_{m-1},\,y_\text{src}\big),\qquad m=1,\dots,M.
\end{equation}
The terminal latent is defined as
\begin{equation}
\label{eq:f_phi_def}
z_T \;\triangleq\; z_{t_M} \;\equiv\; f_\phi(x_0,y_\text{src}),
\end{equation}
which serves as the starting point for the backward (generation/denoising) process. To reconstruct the source, we denoise from $z_T$ under the same prompt $y_\text{src}$. We initialize the backward chain by explicitly setting
\begin{equation}
\label{eq:backward_init}
\hat z_{t_M} \;\triangleq\; z_T,
\end{equation}
and apply $M$ backward steps:
\begin{equation}
\label{eq:backward_step}
\hat z_{t_{m-1}}=G_\theta\!\big(\hat z_{t_m},\,t_m,\,y_\text{src}\big),\qquad m=M,\dots,1.
\end{equation}
We define the induced image-generation mapping of the backward model as
\begin{equation}
\label{eq:g_theta_def}
g_\theta(z_T,y)\;\triangleq\;\mathrm{dec}(\hat z_{t_0}),\quad \text{where }\hat z_{t_0}\text{ is obtained by running \eqref{eq:backward_init}--\eqref{eq:backward_step} under prompt }y.
\end{equation}
Thus, the reconstructed image is
\begin{equation}
\label{eq:x_src_def}
\hat x_\text{src}=g_\theta\!\big(f_\phi(x_0,y_\text{src}),\,y_\text{src}\big),
\end{equation}
and we optimize both adapters end-to-end with the reconstruction loss
\begin{equation}
\label{eq:rec_loss}
\mathcal L_\text{rec}(\phi,\theta)=\mathrm{LPIPS}(\hat x_\text{src},x_0).
\end{equation}
This term enforces faithful inversion and content preservation under the source prompt.

\paragraph{Editing.}
Minimizing $\mathcal L_\text{rec}$ alone admits a degenerate shortcut: the forward model can keep $z_T$ insufficiently noised, so the backward pass reconstructs $x_0$ well but becomes weakly sensitive to $y_\text{trg}$, yielding near-identity outputs. We therefore add complementary terms that (i) keep the terminal latent compatible with the backbone's noise prior, preventing under-noising, and (ii) explicitly encourage prompt-consistent semantic change.

\paragraph{Prior matching for $z_T$.}
We regularize the terminal latent $z_T$ to remain compatible with the backbone's noise
prior $\mathcal{N}(0, I)$, using two complementary terms that constrain different
statistics of $\tilde z = z_T / \sigma_T \in \mathbb{R}^{C\times H\times W}$.

\emph{Channel-space Gaussianity.} Treating the spatial tokens of $\tilde z$ as samples
in $\mathbb{R}^C$, we estimate the empirical mean $\mu(\tilde z) \in \mathbb{R}^C$ and
covariance $\Sigma(\tilde z) \in \mathbb{R}^{C\times C}$, and penalize their KL
divergence to the standard normal:
\begin{equation}
\label{eq:kl_loss}
\mathcal{L}_{\mathrm{KL}}(z_T)
= \mathrm{KL}\!\bigl(\mathcal{N}(\mu, \Sigma)\,\|\,\mathcal{N}(0, I)\bigr)
= \tfrac{1}{2}\bigl(\mathrm{tr}(\Sigma) + \mu^{\top}\mu - C - \log\det\Sigma\bigr).
\end{equation}
This matches the per-channel means and the full channel covariance to the identity,
driving the channels toward unit variance and mutual decorrelation.

\emph{Spatial whiteness.} The channel-space statistics above are invariant to spatial
permutations of $\tilde z$, and therefore cannot detect structured low-frequency content
(smooth, content-bearing residuals) in the terminal latent. We close this blind spot
with a low-frequency excess penalty. Let
$\bar z = \mathrm{AvgPool}_{k\times k}(\tilde z) \in \mathbb{R}^{C\times H/k\times W/k}$
denote the $k\times k$ average-pooled response, with $k = 8$. If $\tilde z$ were i.i.d.\
$\mathcal{N}(0, 1)$, each pooled entry would be the sample mean of $k^2$ unit-variance
values, hence zero-mean with variance $1/k^2$. We penalize energy in excess of this
white-noise floor with a one-sided hinge:
\begin{equation}
\label{eq:lf_loss}
\mathcal{L}_{\mathrm{LF}}(z_T)
= \bigl[\,\mathbb{E}\!\bigl[\bar z^{\,2}\bigr] - 1/k^{2}\,\bigr]_{+},
\qquad [u]_+ = \max(u, 0).
\end{equation}

The two terms are complementary: $\mathcal{L}_{\mathrm{KL}}$ constrains the channel-space
statistics (per-channel mean and full inter-channel covariance), while
$\mathcal{L}_{\mathrm{LF}}$ constrains spatial whiteness at scale $k$. Both vanish in
expectation when $\tilde z \sim \mathcal{N}(0, I)$, so neither produces gradient from
typical i.i.d.\ fluctuations; $\mathcal{L}_{\mathrm{LF}}$ is moreover one-sided,
penalizing only excess low-frequency structure and not a deficit.

\textbf{Directional CLIP for target edits.}
To enforce that the produced change follows the target prompt, we generate an edited image by denoising from the same inverted latent but conditioning on $y_\text{trg}$:
\begin{equation}
\label{eq:x_trg_def}
\hat x_\text{trg}=g_\theta\!\big(f_\phi(x_0,y_\text{src}),\,y_\text{trg}\big).
\end{equation}
Let $E_I(\cdot)$ and $E_T(\cdot)$ denote CLIP image and text encoders. Following~\cite{DBLP:journals/corr/abs-2108-00946,kim2022diffusioncliptextguideddiffusionmodels}, we define the induced image direction $\Delta E_I=E_I(\hat x_\text{trg})-E_I(x_0)$ and the desired text direction $\Delta E_T=E_T(y_\text{trg})-E_T(y_\text{src})$, and maximize their alignment:
\begin{equation}
\label{eq:dclip_loss}
\mathcal L_\text{dCLIP} = -\big\langle \Delta E_I,\Delta E_T\big\rangle.
\end{equation}
Together, $\mathcal L_\text{rec}$, $\mathcal L_\mathrm{KL}$, $\mathcal L_\mathrm{LF}$, and $\mathcal L_\text{dCLIP}$ enforce the key trade-off in editing: preserve source-specific content while producing strong, prompt-consistent modifications.




\paragraph{Final Objective.}
We jointly optimize the forward and backward adapters on triplets $(x_0,\allowbreak y_\text{src},\allowbreak y_\text{trg})$ by combining reconstruction, edit-direction alignment, and terminal-prior regularization. Using the definitions in \eqref{eq:x_src_def} and \eqref{eq:x_trg_def}, the overall objective is
\begin{equation}
\label{eq:final_objective}
\mathcal L(\phi,\theta; x_0,y_\text{src},y_\text{trg})
=
\mathcal L_\text{rec}(\phi,\theta)
+
\mathcal L_\text{dCLIP}
+
\mathcal L_\mathrm{LF}(z_T)
+
\lambda_\mathrm{KL}\,\mathcal L_\mathrm{KL}(z_T)
\end{equation}
where $\mathcal L_\text{rec}$ is given in \eqref{eq:rec_loss}, $\mathcal L_\mathrm{KL}$ in \eqref{eq:kl_loss}, $\mathcal L_\mathrm{LF}$ in \eqref{eq:lf_loss}, $\mathcal L_\text{dCLIP}$ in \eqref{eq:dclip_loss}, and $z_T=f_\phi(x_0,y_\text{src})$ in \eqref{eq:f_phi_def}. Each term is necessary
(Table~\ref{tab:ablation}; full analysis in Appendix~\ref{app:ablations}): reconstruction
alone collapses to identity, the prior terms restore prompt sensitivity, and the
directional signal is essential; training only $F_\phi$ or $G_\theta$ is likewise unstable.

\subsection{Editing inference}
\label{sec:editing_inference}
At test time, editing requires only the learned forward and backward adapters and does
not rely on additional attention-control modules or per-image optimization. Given an
input image $x_0$ with a source prompt $y_\text{src}$ and a target prompt $y_\text{trg}$,
we first compute the terminal latent
\begin{equation}
\label{eq:inference_zT}
z_T = f_\phi(x_0,y_\text{src}),
\end{equation}
using the $M$-step forward process in Eq.~\eqref{eq:forward_step}. We then denoise from
the same $z_T$ under the target prompt and decode with the VAE. To amplify the edit, we
replace the unconditional branch of classifier-free guidance with the
$y_\text{src}$-conditioned prediction, following Delta Denoising Score~\citep{hertz2023deltadenoisingscore}, giving
the guided backward step
\begin{equation}
\label{eq:guided_backward}
\hat z_{t_{m-1}} = G_\theta\!\big(\hat z_{t_m}, t_m, y_\text{src}\big)
+ \omega\,\Big(G_\theta\!\big(\hat z_{t_m}, t_m, y_\text{trg}\big)
- G_\theta\!\big(\hat z_{t_m}, t_m, y_\text{src}\big)\Big),
\qquad m = M,\dots,1,
\end{equation}
controlled by a guidance scale $\omega$. Like our directional CLIP objective
\eqref{eq:dclip_loss}, this update is driven by the
source-to-target difference rather than the absolute target, keeping training and
inference consistent in optimizing \emph{relative change}. We denote the induced guided generation mapping
by $\hat g_\theta(z_T, y, \omega)$; the original $g_\theta$ in Eq.~\eqref{eq:g_theta_def}
is recovered as the special case $\omega = 1$. The final edited image is
\begin{equation}
\label{eq:editing_inference}
\hat x_\text{trg} = \hat g_\theta\!\big(f_\phi(x_0,y_\text{src}),\,y_\text{trg},\,\omega\big).
\end{equation}
This inference pipeline is fully deterministic and performs exactly
$2M$ model calls (forward + backward), making it efficient for
accelerated backbones; output diversity, when desired, is controlled
via~$\omega$ (Appendix~\ref{app:additional_editing}). Crucially, because
$F_\phi$ and $G_\theta$ were jointly trained through the full loop
with the objective in Eq.~(21), the produced $z_T$ is both
prior-compatible and informative enough to preserve content, enabling
strong edits without extra components.

\begin{table}[t]
\centering
\small
\caption{Ablation of objective terms on Z-Image-Turbo (step 600);
bold marks the best \emph{editing run}. Without
$\mathcal{L}_{\text{dCLIP}}$ the model merely reconstructs the source
(dCLIP $0.046$, IR $<0$); the full objective is best on editability
and quality. LPIPS (VGG), SSIM, and PSNR are whole-image.}
\setlength{\tabcolsep}{4pt}
\begin{tabular}{l
  S[table-format=2.2]
  S[table-format=1.4]
  S[table-format=1.4]
  S[table-format=1.4]
  S[table-format=2.2]
  S[table-format=-1.4]}
\toprule
Method & {CLIP$_{\text{wh}}$\,$\uparrow$} & {dCLIP\,$\uparrow$} & {LPIPS\,$\downarrow$} & {SSIM\,$\uparrow$} & {PSNR\,$\uparrow$} & {IR\,$\uparrow$} \\
\midrule
w/o $\mathcal{L}_{\text{KL}}$ & 26.06 & 0.1350 & \textbf{0.1901} & \textbf{0.8306} & 21.81 & 0.6029 \\
w/o $\mathcal{L}_{\text{LF}}$ & 26.13 & 0.1423 & 0.2071 & 0.8171 & 21.49 & 0.6015 \\
\textbf{Ours (full)} & \textbf{26.44} & \textbf{0.1541} & 0.1959 & 0.8296 & \textbf{22.01} & \textbf{0.7383} \\
\midrule
\textit{w/o $\mathcal{L}_{\text{dCLIP}}$ (no edit signal)} & 23.87 & 0.0457 & 0.0761 & 0.9075 & 25.72 & -0.1320 \\
\bottomrule
\end{tabular}
\label{tab:ablation}
\end{table}

\begin{figure}[t]
\centering
\includegraphics[width=0.7\linewidth]{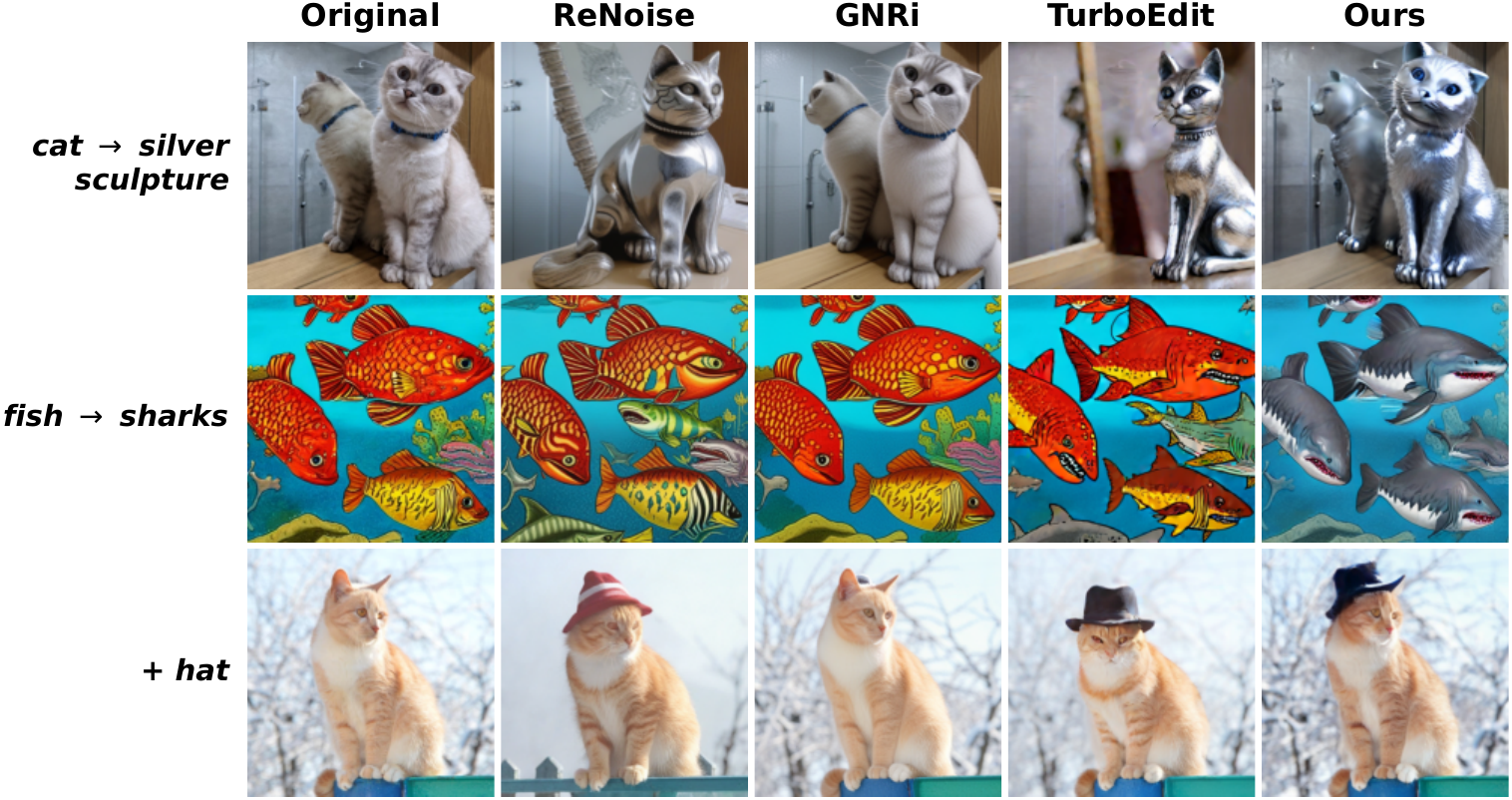}
    \caption{\textbf{Qualitative comparison on SDXL-Turbo} against ReNoise, GNRi, and
TurboEdit (one edit per row). The baselines either lose input identity and detail
or fail to apply the edit; our method applies each edit while preserving
composition, identity, and background.}
\label{fig:sdxl_qual}
    \label{fig:image_edits_examples_sdxl}
\end{figure}
\begin{figure}[t]
\centering
\includegraphics[width=\linewidth]{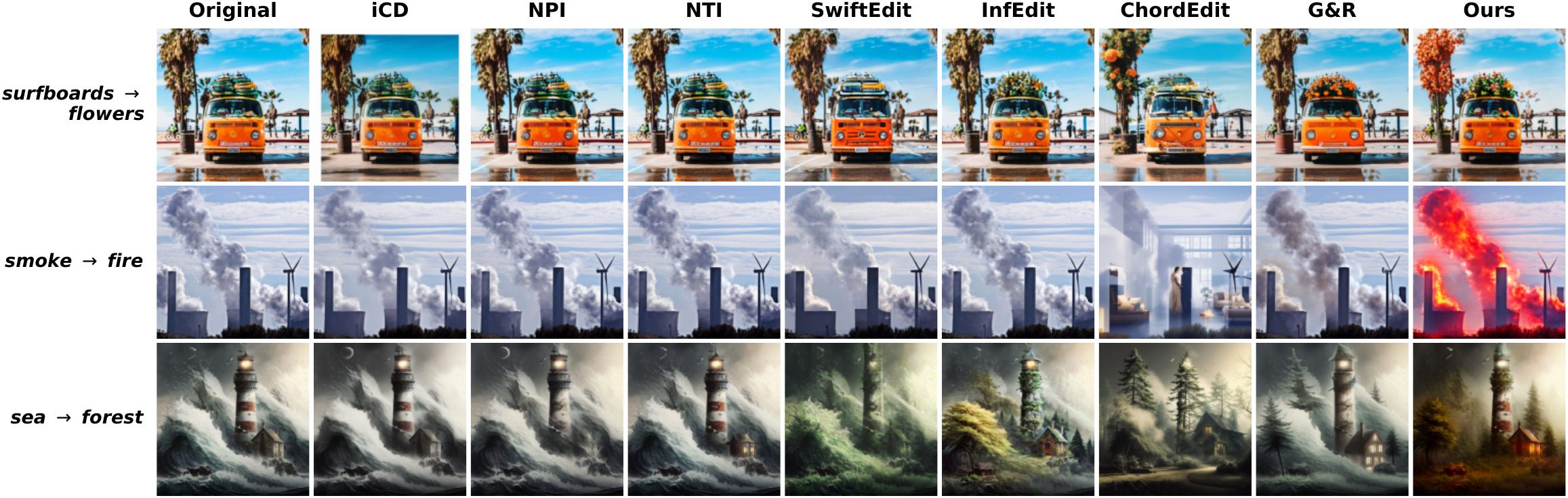}
    \caption{\textbf{Qualitative comparison on LCM-DreamShaper} against iCD, NPI, NTI,
SwiftEdit, InfEdit, ChordEdit, and G\&R (one edit per row). Most baselines barely
edit the input or distort the scene when they do; our method produces a consistent
edit per row while preserving layout, identity, and background.}
    \label{fig:image_edits_examples_sd15}
\end{figure}
\section{Experiments}

\definecolor{TopOne}{HTML}{C6EFCE}    
\definecolor{TopTwo}{HTML}{FFF2CC}    
\definecolor{TopThree}{HTML}{FCE4D6}  
\definecolor{Hdr}{HTML}{F2F2F2}       

%

\definecolor{TopOne}{HTML}{C6EFCE}    
\definecolor{TopTwo}{HTML}{FFF2CC}    
\definecolor{TopThree}{HTML}{FCE4D6}  
\definecolor{Hdr}{HTML}{F2F2F2}       

\begin{table*}[t]
\centering
\small
\setlength{\tabcolsep}{4.2pt}
\renewcommand{\arraystretch}{1.12}
\caption{Baselines vs Ours under the official PIE-Bench protocol
(CLIP\textsubscript{wh}, background-masked CLIP\textsubscript{ed}/LPIPS/SSIM/PSNR and
Structure Distance $\times 10^3$); dCLIP is the directional CLIP score; ImageReward and time. Within
each block, best is \textbf{bold} and second-best is \underline{underlined}.
Avg Rank is the mean dense-rank over the nine metrics, lower is better, with
ties resolved on displayed precision.}
\label{tab:main_results}
\resizebox{\textwidth}{!}{%
\begin{tabular}{l
S[table-format=2.2]  
S[table-format=2.2]  
S[table-format=1.3]  
S[table-format=2.1]  
S[table-format=1.3]  
S[table-format=1.3]  
S[table-format=2.1]  
S[table-format=1.3,table-space-text-pre=-] 
S[table-format=3.1]  
S[table-format=1.2]  
}
\toprule
 & \multicolumn{3}{c}{Editability} & \multicolumn{4}{c}{Preservation} & \multicolumn{1}{c}{Quality} & \multicolumn{1}{c}{Efficiency} & \multicolumn{1}{c}{Summary} \\
\cmidrule(lr){2-4}\cmidrule(lr){5-8}\cmidrule(lr){9-9}\cmidrule(lr){10-10}\cmidrule(lr){11-11}
Method &
\multicolumn{1}{c}{CLIP\textsubscript{wh}$\uparrow$} &
\multicolumn{1}{c}{CLIP\textsubscript{ed}$\uparrow$} &
\multicolumn{1}{c}{dCLIP$\uparrow$} &
\multicolumn{1}{c}{Struct$\downarrow$} &
\multicolumn{1}{c}{LPIPS$\downarrow$} &
\multicolumn{1}{c}{SSIM$\uparrow$} &
\multicolumn{1}{c}{PSNR$\uparrow$} &
\multicolumn{1}{c}{IR$\uparrow$} &
\multicolumn{1}{c}{Time$\downarrow$ (s)} &
\multicolumn{1}{c}{Avg Rank$\downarrow$} \\
\midrule
\multicolumn{11}{l}{\textit{SD-based models}}\\
\midrule
G\&R      & \textbf{26.47} & \textbf{22.79} & 0.106 & \underline{13.7} & 0.070 & 0.834 & 25.0 & 0.427 & 19.7 & \textbf{3.33} \\
InfEdit   & 25.57 & 22.24 & 0.096 & 18.5 & \underline{0.053} & \underline{0.850} & \textbf{28.0} & 0.278 & 3.5 & \textbf{3.33} \\
ChordEdit & 25.90 & \underline{22.54} & \underline{0.119} & 29.7 & 0.122 & 0.764 & 22.4 & \underline{0.516} & 3.5 & 5.11 \\
SwiftEdit & 24.61 & 21.34 & 0.081 & 16.2 & 0.077 & 0.820 & 24.0 & -0.104 & \textbf{0.2} & 5.78 \\
iCD       & 24.93 & 21.71 & 0.047 & 21.4 & 0.084 & 0.820 & 24.8 & -0.061 & 1.1 & 5.67 \\
NPI       & 24.81 & 21.57 & 0.085 & 13.8 & 0.062 & 0.843 & 26.7 & 0.007 & 9.1 & 4.56 \\
NTI       & 24.35 & 21.11 & 0.050 & \textbf{9.6} & \textbf{0.051} & \textbf{0.853} & \underline{27.9} & -0.093 & 115.0 & 4.78 \\
\addlinespace[1pt]
\cmidrule(lr){1-11}
\addlinespace[1pt]
\textbf{Ours} & \underline{26.13} & 22.47 & \textbf{0.155} & 22.8 & 0.071 & 0.843 & 23.6 & \textbf{0.573} & \underline{0.6} & \underline{3.44} \\
\midrule
\multicolumn{11}{l}{\textit{SDXL-based models}}\\
\midrule
TurboEdit & \underline{26.52} & \underline{23.05} & \underline{0.145} & \underline{30.4} & \underline{0.102} & 0.801 & \underline{22.5} & \underline{0.670} & \underline{1.0} & \underline{2.22} \\
GNRi      & 24.12 & 20.74 & 0.038 & \textbf{8.7} & \textbf{0.064} & \textbf{0.837} & \textbf{27.7} & -0.129 & \textbf{0.6} & 2.33 \\
ReNoise   & 25.36 & 22.21 & 0.124 & 61.0 & 0.185 & 0.714 & 20.3 & 0.324 & 3.1 & 3.56 \\
\addlinespace[1pt]
\cmidrule(lr){1-11}
\addlinespace[1pt]
\textbf{Ours} & \textbf{27.17} & \textbf{23.31} & \textbf{0.178} & 43.0 & 0.105 & \underline{0.808} & 20.8 & \textbf{0.858} & \textbf{0.6} & \textbf{1.78} \\
\midrule
\multicolumn{11}{l}{\textit{Flow-based models}}\\
\midrule
RF-Edit      & 25.55 & 22.44 & 0.103 & 21.3 & 0.104 & 0.855 & 25.0 & 0.449 & 30.0 & 3.89 \\
FlowEdit     & \underline{26.17} & \textbf{22.90} & 0.108 & 28.4 & 0.109 & 0.841 & 22.0 & \underline{0.726} & \underline{8.0} & 3.22 \\
FireFlow     & 26.06 & 22.76 & 0.117 & 26.4 & 0.119 & 0.830 & 23.2 & 0.626 & 21.0 & 3.78 \\
UniEdit-Flow & 25.64 & 22.22 & \underline{0.123} & \textbf{9.3} & \underline{0.055} & \textbf{0.917} & \textbf{29.7} & 0.360 & 10.0 & \underline{2.67} \\
\addlinespace[1pt]
\cmidrule(lr){1-11}
\addlinespace[1pt]
\textbf{Ours} 
& \textbf{26.43} & \underline{22.77} & \textbf{0.154} & \underline{12.8} & \textbf{0.051} & \underline{0.904} & \underline{26.2} & \textbf{0.738} & \textbf{3.2} & \textbf{1.44} \\
\bottomrule
\end{tabular}%
}
\end{table*}


\paragraph{Backbones and evaluation protocol.}
We evaluate our framework across three families of accelerated text-to-image
backbones and compare methods within the corresponding family. For the
\textbf{SD family}, we include both full-step SD editors (e.g., NTI, NPI,
G\&R) and distilled SD variants used by accelerated editors (e.g.,
DreamShaper-LCM for InfEdit and our method, and the official iCD distilled
model). For the \textbf{SDXL family}, all compared accelerated editors
(TurboEdit, ReNoise, GNRi, and ours) are run on \textbf{SDXL-Turbo}. For the
\textbf{flow / rectified-flow family}, we use \textbf{Z-Image Turbo} as the
distilled flow backbone and compare against RF-based baselines RF-Edit,
FlowEdit, FireFlow, and UniEdit-Flow under the same setting. This split matches
our qualitative figures: Fig.~\ref{fig:image_edits_examples_sd15} (DreamShaper-LCM),
Fig.~\ref{fig:image_edits_examples_sdxl} (SDXL-Turbo), and Fig.~\ref{fig:image_edits_examples_flow}
(Z-Image Turbo).

\paragraph{Fine-tune setup.}
For each backbone family (DreamShaper-LCM / SDXL-Turbo / Z-Image Turbo), we fine-tune two LoRA adapters (forward $F_\phi$ and backward $G_\theta$) by minimizing the full objective in Eq.~\ref{eq:final_objective}.
The editing term uses the directional CLIP loss (Eq.~\ref{eq:dclip_loss}), aligning the image embedding shift $E_I(\hat x_{\text{trg}})-E_I(x_0)$ with the text shift $E_T(y_{\text{trg}})-E_T(y_{\text{src}})$, while the reconstruction term uses LPIPS (Eq.~\ref{eq:rec_loss}, VGG-16).
CLIP embeddings are computed with CLIP ViT-L/14 on $224{\times}224$ resized images.
We fine-tune $(\phi,\theta)$ on curated triplets from HQ-Edit~\citep{hui2024hqedithighqualitydatasetinstructionbased}, using
$300$ iterations (batch size $16$) for SD1.5, $600$ iterations (batch size $16$) for
SDXL-Turbo, and $600$ iterations (batch size $4$) for Z-Image-Turbo.
Additional optimizer and hyperparameter details are provided in Appendix~\ref{sec:fine-tune_setup}.
\begin{table}
\centering
\caption{\textbf{Human preference study on Z-Image-Turbo (flow-based).} Pairwise
comparisons of our method against each flow editor, $50$ per baseline ($200$
total). Each cell reports the preference score $\mathrm{win}+\tfrac{1}{2}\mathrm{tie}$
(\%); $50\%$ is parity, higher favors ours. \emph{Overall} is overall preference,
\emph{Match} is agreement with the edit prompt, and \emph{Preserve} is fidelity to
the source image.}
\label{tab:human}
\small
\begin{tabular*}{\textwidth}{@{\extracolsep{\fill}}lccc@{}}
\toprule
Ours vs. & Overall $\uparrow$ & Match $\uparrow$ & Preserve $\uparrow$ \\
\midrule
RF-Edit      & 78.0 & 59.0 & 81.0 \\
FlowEdit     & 76.0 & 58.0 & 81.0 \\
FireFlow     & 81.0 & 58.0 & 83.0 \\
UniEdit-Flow & 70.0 & 59.0 & 70.0 \\
\midrule
\textbf{Aggregate} & \textbf{76.3} & \textbf{58.5} & \textbf{78.8} \\
\bottomrule
\end{tabular*}
\end{table}
\paragraph{Inversion and editing setup.}
We evaluate inversion and editing on separate datasets. For the inversion
experiments, we use more than $2700$ high-resolution images from
MS-COCO~\cite{lin2015microsoftcococommonobjects} for quantitative evaluation, as it provides
a large amount of labeled data at varying scales, enabling reliable detection of
differences between methods. We report reconstruction only for methods to which
inversion is applicable, since inversion-free editors do not recover a source
latent and are therefore not comparable on this task; for these reconstruction
comparisons we disable classifier-free guidance for all methods. For the editing experiments, we evaluate on the full PIE-Bench
benchmark of $700$ images spanning its ten edit categories, with all
metrics computed by the official PIE-Bench
implementation~\cite{ju2023directinversionboostingdiffusionbased}. Our method uses neither blend words nor
any additional optimization at inference, and all baselines are run with their
default settings as provided by the authors or official implementations. 
\subsection{Image Inversion}
We assess inversion through image reconstruction, comparing within each backbone
family against methods for which inversion is applicable: DDIM
inversion~\citep{song2022denoisingdiffusionimplicitmodels} and
iCD~\citep{starodubcev2024invertibleconsistencydistillationtextguided} for SD,
SwiftEdit~\citep{Nguyen_2025_CVPR} as a one-step inversion baseline,
GNRi~\citep{samuel2025lightningfastimageinversionediting},
TurboEdit~\citep{deutch2024turboedittextbasedimageediting}, and
ReNoise~\citep{garibi2024renoiserealimageinversion} for SDXL-Turbo, and the
flow-based methods UniEdit-Flow~\citep{jiao2025unieditflowunleashinginversionediting},
FireFlow~\citep{deng2025fireflow}, and
RF-Edit~\citep{wang2025tamingrectifiedflowinversion}. We measure reconstruction
quality with mean-squared error (MSE), ImageReward, and LPIPS.
 
As reported in Table~\ref{tab:reconstruction_results_horizontal} (qualitative reconstructions and per-backbone figures in Appendix~\ref{app:additional_inversion}),
our method attains the lowest MSE and LPIPS in every backbone family, indicating
the most faithful pixel- and perceptual-level reconstruction; on the SD
backbone it surpasses even full-step DDIM inversion despite operating in the
accelerated few-step regime. It also achieves the best ImageReward on the SDXL
and flow backbones. The only exception is ImageReward on SD, where SwiftEdit is marginally higher (0.437 vs. 0.425) at far lower fidelity (LPIPS 0.507 vs. 0.172, MSE 0.0496 vs. 0.0056). Overall, our adapters recover the source
latent more accurately than prior accelerated inversion methods while matching or
exceeding their perceptual quality.




\begin{table}[t]
\centering
\caption{Evaluation metrics for image reconstruction across SD-, SDXL-, and flow-based methods. Best (\textbf{bold}) and second best (\underline{underlined}) are computed within each block for each metric. Only methods that perform an explicit inversion are included.}
\label{tab:reconstruction_results_horizontal}
\small
\setlength{\tabcolsep}{4pt} 
\resizebox{\linewidth}{!}{%
\begin{tabular}{@{}lrrr @{\hskip 3.0em} lrrr @{\hskip 3.0em} lrrr@{}}
\toprule
\multicolumn{4}{c}{\textit{SD-based methods}} &
\multicolumn{4}{c}{\textit{SDXL-based methods}} &
\multicolumn{4}{c}{\textit{Flow-based methods}} \\
\cmidrule(l{4pt}r{4pt}){1-4}\cmidrule(l{4pt}r{4pt}){5-8}\cmidrule(l{4pt}r{4pt}){9-12}
Model & MSE $\downarrow$ & ImageReward $\uparrow$ & LPIPS $\downarrow$ &
Model & MSE $\downarrow$ & ImageReward $\uparrow$ & LPIPS $\downarrow$ &
Model & MSE $\downarrow$ & ImageReward $\uparrow$ & LPIPS $\downarrow$ \\
\cmidrule(l{4pt}r{4pt}){1-4}\cmidrule(l{4pt}r{4pt}){5-8}\cmidrule(l{4pt}r{4pt}){9-12}
DDIM-50 & \underline{0.0079} & -0.023 & \underline{0.279} &
GNRi & \underline{0.0065} & 0.416 & \underline{0.283} &
UniEdit-Flow & 0.0127 & \underline{-0.050} & \underline{0.274} \\
iCD & 0.0447 & -0.497 & 0.611 &
TurboEdit & 0.0470 & \underline{0.427} & 0.483 &
FireFlow & \underline{0.0092} & -0.102 & 0.327 \\
SwiftEdit & 0.0496 & \textbf{0.437} & 0.507 &
ReNoise & 0.0615 & 0.367 & 0.541 &
RF-Edit & 0.0119 & -0.342 & 0.376 \\
\midrule
\textbf{Ours} & \textbf{0.0056} & \underline{0.425} & \textbf{0.172} &
\textbf{Ours} & \textbf{0.0061} & \textbf{0.504} & \textbf{0.186} &
\textbf{Ours} & \textbf{0.0052} & \textbf{0.535} & \textbf{0.166} \\
\bottomrule
\end{tabular}%
}
\end{table}

\subsection{Text-guided image editing}
We compare within each backbone family, matching
Table~\ref{tab:main_results}: for SD, full-step editors
(NTI~\cite{Mokady_2023_CVPR},
NPI~\cite{miyake2024negativepromptinversionfastimage},
G\&R~\cite{titov2024guideandrescaleselfguidancemechanismeffective})
and accelerated ones
(iCD~\cite{starodubcev2024invertibleconsistencydistillationtextguided},
InfEdit~\cite{xu2023inversionfreeimageeditingnatural},
ChordEdit~\cite{lu2026chordedit}, SwiftEdit~\cite{Nguyen_2025_CVPR});
for SDXL-Turbo,
TurboEdit~\cite{deutch2024turboedittextbasedimageediting},
ReNoise~\cite{garibi2024renoiserealimageinversion}, and
GNRi~\cite{samuel2025lightningfastimageinversionediting}; and for the
flow family, RF-Edit~\cite{wang2025tamingrectifiedflowinversion},
FireFlow~\cite{deng2025fireflow},
UniEdit-Flow~\cite{jiao2025unieditflowunleashinginversionediting}, and
FlowEdit~\cite{kulikov2025floweditinversionfreetextbasedediting}.
\paragraph{Qualitative evaluation.}
We show representative results for each backbone family: LCM-DreamShaper~\citep{luo2023latentconsistencymodelssynthesizing}
in Figure~\ref{fig:image_edits_examples_sd15}, SDXL-Turbo~\citep{sauer2023adversarialdiffusiondistillation}
in Figure~\ref{fig:image_edits_examples_sdxl}, and the flow-based Z-Image
Turbo~\citep{zimage2025} in Figure~\ref{fig:image_edits_examples_flow}. Across all
three, our method applies the requested edit while preserving the layout,
identity, and background of the input. The baselines instead exhibit two recurring failure modes. Many
under-edit: on SD, iCD, NPI, NTI, SwiftEdit, and InfEdit frequently
produce little visible change (e.g.,
\emph{surfboards$\rightarrow$flowers}, \emph{smoke$\rightarrow$fire}).
Those that do edit tend to over-edit: ChordEdit distorts the
surrounding scene, ReNoise and TurboEdit sacrifice pose, background,
and identity, and the flow baselines drift out of the input domain. In
contrast, our method produces a consistent, visible edit in every
case; further per-backbone examples are in
Appendix~\ref{app:additional_editing}.

\textbf{Quantitative evaluation.} We group metrics into four aspects,
editability, preservation, quality, and efficiency, and summarize
overall standing with the mean rank across metrics (Avg Rank);
per-family results are in Table~\ref{tab:main_results}.

\textbf{Editability} is measured by the masked CLIP (CLIP ViT-L/14) similarities
(CLIP\textsubscript{wh}, CLIP\textsubscript{ed}) and the directional
dCLIP score (Eq.~\ref{eq:dclip_loss}, additionally measured with OpenCLIP ViT-bigG/14 in Appendix~\ref{app:heldout_clip}), reflecting how well the edited image follows the target
instruction. Our method achieves the top dCLIP in every block and the
top CLIP\textsubscript{wh} in the SDXL and flow blocks, remaining a
close second in the SD block. The dCLIP values also quantify the
under-editing seen qualitatively in
Figures~\ref{fig:image_edits_examples_flow},
\ref{fig:image_edits_examples_sdxl},
and~\ref{fig:image_edits_examples_sd15}:
many SD baselines apply very weak edits (e.g., $0.047$ for iCD,
$0.050$ for NTI, $0.081$ for SwiftEdit, against our $0.155$),
consistent with the figures where these methods leave the source
nearly unchanged.

\textbf{Preservation} is assessed with Structure Distance, LPIPS (SqueezeNet),
SSIM, and PSNR, capturing fidelity to the input $x_0$ in non-edited
regions. Our method attains the best LPIPS in the flow block and stays
competitive across all families. Where a baseline edges ahead on
preservation, it does so at far lower edit strength (GNRi on SDXL:
dCLIP $0.038$) or higher inference cost (UniEdit-Flow on flow),
reflecting a conservative operating point rather than a better
trade-off; consistently, Structure Distance favours conservative
editors, and its per-block winners (NTI, GNRi, UniEdit-Flow) edit
near least. Sweeping the guidance scale $\omega$ traces a full
quality--preservation frontier that dominates the baselines' single
operating points (Appendix~\ref{app:additional_editing}).\looseness=-1

\textbf{Quality} is measured by Image Reward, a learned proxy for human preference.
Our method attains the best Image Reward in every block. Notably, several
under-editing baselines score at or below zero (e.g., iCD, NTI on SD;
GNRi on SDXL), confirming that producing little change is penalised as low-quality
rather than rewarded as faithful preservation.\looseness=-1

\textbf{Efficiency.} Our method yields the lowest or near-lowest
inference time in each block, several times faster than the competing
flow editors. It attains the best Avg Rank on the SDXL and flow
backbones; on SD it ranks a close third (3.44 vs.\ 3.33) behind G\&R and InfEdit, which are 6--30+$\times$ slower. Times are inference only: ours, iCD and SwiftEdit train once up front, the rest do not (Appendix~\ref{sec:fine-tune_setup}).\looseness=-1

\textbf{Human evaluation.} A pairwise study on Z-Image-Turbo against
the four flow editors (Appendix~\ref{app:human}, Table~\ref{tab:human})
confirms the metrics: ours is preferred on overall quality ($76.3\%$)
and preservation ($78.8\%$) and at near parity on prompt match ($58.5\%$),
i.e.\ it applies the same edits while preserving more of the input.\looseness=-1
\section{Conclusion}
We introduced a simple, universal framework for text-guided editing with accelerated
backbones: inversion and generation form a single differentiable graph, and we optimize
a task-level objective over the full trajectory rather than per-step surrogates.
Fine-tuning two lightweight LoRA adapters (forward and backward) jointly enforces
faithful reconstruction, prompt-consistent change, and a well-posed terminal latent.
Across consistency-distilled, generator-based, and flow-based
backbones, it attains the strongest edit alignment in each family,
with preservation that matches or exceeds prior editors at matched
edit strength, efficient inference, and no backbone-specific inversion
or test-time optimization. Future work includes
instruction-style edits, higher resolutions, and video editing. 

\section{Acknowledgment} 

This work was supported by the Ministry of Economic Development of the Russian Federation (Agreement No.~\mbox{139-10-2025-034} dated June 19, 2025, unique identifier IGK 000000\allowbreak C313925P4D0002).
\bibliography{egbib}

@String(CVPR  = {IEEE Conf. Comput. Vis. Pattern Recog.})

@String(ICCV  = {Int. Conf. Comput. Vis.})

@String(ECCV  = {Eur. Conf. Comput. Vis.})

@String(NeurIPS = {Adv. Neural Inform. Process. Syst.})

@String(ICML  = {Int. Conf. Mach. Learn.})

@String(ICLR  = {Int. Conf. Learn. Represent.})

@String(CVPR  = {CVPR})

@String(ICCV  = {ICCV})

@String(ECCV  = {ECCV})

@String(NeurIPS = {NeurIPS})

@String(ICML  = {ICML})

@String(ICLR  = {ICLR})

@article{Wu_2026,
  author       = {Tao Wu and
                  Senmao Li and
                  Yaxing Wang and
                  Shiqi Yang and
                  Kai Wang and
                  Joost van de Weijer},
  title        = {Training-free image inversion for one-step diffusion models},
  journal      = {Pattern Recognit.},
  volume       = {180},
  pages        = {114063},
  year         = {2026},
  url          = {https://doi.org/10.1016/j.patcog.2026.114063},
  doi          = {10.1016/J.PATCOG.2026.114063},
  bibsource    = {dblp computer science bibliography, https://dblp.org}
}

@article{dai2026bifmbidirectionalflowmatching,
  author       = {Yasong Dai and
                  Zeeshan Hayder and
                  David Ahmedt{-}Aristizabal and
                  Hongdong Li},
  title        = {BiFM: Bidirectional Flow Matching for Few-Step Image Editing and Generation},
  journal      = {CoRR},
  volume       = {abs/2603.24942},
  year         = {2026},
  url          = {https://doi.org/10.48550/arXiv.2603.24942},
  doi          = {10.48550/ARXIV.2603.24942},
  eprinttype   = {arXiv},
  eprint       = {2603.24942},
  bibsource    = {dblp computer science bibliography, https://dblp.org}
}

@inproceedings{zuo2025precise,
  author       = {Jing Zuo and
                  Luoping Cui and
                  Chuang Zhu and
                  Yonggang Qi},
  editor       = {Danielle Belgrave and
                  Cheng Zhang and
                  Laura N. Montoya and
                  Hsuan{-}Tien Lin and
                  Razvan Pascanu and
                  Piotr Koniusz and
                  Marzyeh Ghassemi and
                  Nancy Chen and
                  Iv{\'{a}}n Vladimir Meza Ru{\'{\i}}z and
                  Arturo Loaiza{-}Bonilla},
  title        = {Precise Diffusion Inversion: Towards Novel Samples and Few-Step Models},
  booktitle    = {Advances in Neural Information Processing Systems 38: Annual Conference
                  on Neural Information Processing Systems 2025, NeurIPS 2025, San Diego,
                  CA, USA, December 2-7, 2025 / Mexico City, Mexico, November 30 - December
                  5, 2025},
  year         = {2025},
  url          = {http://papers.nips.cc/paper\_files/paper/2025/hash/2e4bc9f6e31aa27861299940aa5242ad-Abstract-Conference.html},
  bibsource    = {dblp computer science bibliography, https://dblp.org}
}

@inproceedings{cao2023masactrltuningfreemutualselfattention,
  author       = {Mingdeng Cao and
                  Xintao Wang and
                  Zhongang Qi and
                  Ying Shan and
                  Xiaohu Qie and
                  Yinqiang Zheng},
  title        = {MasaCtrl: Tuning-Free Mutual Self-Attention Control for Consistent
                  Image Synthesis and Editing},
  booktitle    = {{IEEE/CVF} International Conference on Computer Vision, {ICCV} 2023,
                  Paris, France, October 1-6, 2023},
  pages        = {22503--22513},
  publisher    = {{IEEE}},
  year         = {2023},
  url          = {https://doi.org/10.1109/ICCV51070.2023.02062},
  doi          = {10.1109/ICCV51070.2023.02062},
  bibsource    = {dblp computer science bibliography, https://dblp.org}
}

@inproceedings{hertz2022prompttopromptimageeditingcross,
  author       = {Amir Hertz and
                  Ron Mokady and
                  Jay Tenenbaum and
                  Kfir Aberman and
                  Yael Pritch and
                  Daniel Cohen{-}Or},
  title        = {Prompt-to-Prompt Image Editing with Cross-Attention Control},
  booktitle    = {The Eleventh International Conference on Learning Representations,
                  {ICLR} 2023, Kigali, Rwanda, May 1-5, 2023},
  publisher    = {OpenReview.net},
  year         = {2023},
  url          = {https://openreview.net/forum?id=\_CDixzkzeyb},
  bibsource    = {dblp computer science bibliography, https://dblp.org}
}

@inproceedings{sauer2023adversarialdiffusiondistillation,
  author       = {Axel Sauer and
                  Dominik Lorenz and
                  Andreas Blattmann and
                  Robin Rombach},
  editor       = {Ales Leonardis and
                  Elisa Ricci and
                  Stefan Roth and
                  Olga Russakovsky and
                  Torsten Sattler and
                  G{\"{u}}l Varol},
  title        = {Adversarial Diffusion Distillation},
  booktitle    = {Computer Vision - {ECCV} 2024 - 18th European Conference, Milan, Italy,
                  September 29-October 4, 2024, Proceedings, Part {LXXXVI}},
  series       = {Lecture Notes in Computer Science},
  volume       = {15144},
  pages        = {87--103},
  publisher    = {Springer},
  year         = {2024},
  url          = {https://doi.org/10.1007/978-3-031-73016-0\_6},
  doi          = {10.1007/978-3-031-73016-0\_6},
  bibsource    = {dblp computer science bibliography, https://dblp.org}
}

@inproceedings{titov2024guideandrescaleselfguidancemechanismeffective,
  author       = {Vadim Titov and
                  Madina Khalmatova and
                  Alexandra Ivanova and
                  Dmitry P. Vetrov and
                  Aibek Alanov},
  editor       = {Ales Leonardis and
                  Elisa Ricci and
                  Stefan Roth and
                  Olga Russakovsky and
                  Torsten Sattler and
                  G{\"{u}}l Varol},
  title        = {Guide-and-Rescale: Self-guidance Mechanism for Effective Tuning-Free
                  Real Image Editing},
  booktitle    = {Computer Vision - {ECCV} 2024 - 18th European Conference, Milan, Italy,
                  September 29-October 4, 2024, Proceedings, Part {LXXI}},
  series       = {Lecture Notes in Computer Science},
  volume       = {15129},
  pages        = {235--251},
  publisher    = {Springer},
  year         = {2024},
  url          = {https://doi.org/10.1007/978-3-031-73209-6\_14},
  doi          = {10.1007/978-3-031-73209-6\_14},
  bibsource    = {dblp computer science bibliography, https://dblp.org}
}

@inproceedings{song2023consistencymodels,
  author       = {Yang Song and
                  Prafulla Dhariwal and
                  Mark Chen and
                  Ilya Sutskever},
  editor       = {Andreas Krause and
                  Emma Brunskill and
                  Kyunghyun Cho and
                  Barbara Engelhardt and
                  Sivan Sabato and
                  Jonathan Scarlett},
  title        = {Consistency Models},
  booktitle    = {International Conference on Machine Learning, {ICML} 2023, 23-29 July
                  2023, Honolulu, Hawaii, {USA}},
  series       = {Proceedings of Machine Learning Research},
  volume       = {202},
  pages        = {32211--32252},
  publisher    = {{PMLR}},
  year         = {2023},
  url          = {https://proceedings.mlr.press/v202/song23a.html},
  bibsource    = {dblp computer science bibliography, https://dblp.org}
}

@inproceedings{sauer2024fasthighresolutionimagesynthesis,
  author       = {Axel Sauer and
                  Frederic Boesel and
                  Tim Dockhorn and
                  Andreas Blattmann and
                  Patrick Esser and
                  Robin Rombach},
  editor       = {Takeo Igarashi and
                  Ariel Shamir and
                  Hao (Richard) Zhang},
  title        = {Fast High-Resolution Image Synthesis with Latent Adversarial Diffusion
                  Distillation},
  booktitle    = {{SIGGRAPH} Asia 2024 Conference Papers, {SA} 2024, Tokyo, Japan, December
                  3-6, 2024},
  pages        = {106:1--106:11},
  publisher    = {{ACM}},
  year         = {2024},
  url          = {https://doi.org/10.1145/3680528.3687625},
  doi          = {10.1145/3680528.3687625},
  bibsource    = {dblp computer science bibliography, https://dblp.org}
}

@inproceedings{kim2022diffusioncliptextguideddiffusionmodels,
  author       = {Gwanghyun Kim and
                  Taesung Kwon and
                  Jong Chul Ye},
  title        = {DiffusionCLIP: Text-Guided Diffusion Models for Robust Image Manipulation},
  booktitle    = {{IEEE/CVF} Conference on Computer Vision and Pattern Recognition,
                  {CVPR} 2022, New Orleans, LA, USA, June 18-24, 2022},
  pages        = {2416--2425},
  publisher    = {{IEEE}},
  year         = {2022},
  url          = {https://doi.org/10.1109/CVPR52688.2022.00246},
  doi          = {10.1109/CVPR52688.2022.00246},
  bibsource    = {dblp computer science bibliography, https://dblp.org}
}

@article{DBLP:journals/corr/abs-2108-00946,
  author       = {Rinon Gal and
                  Or Patashnik and
                  Haggai Maron and
                  Amit H. Bermano and
                  Gal Chechik and
                  Daniel Cohen{-}Or},
  title        = {StyleGAN-NADA: CLIP-guided domain adaptation of image generators},
  journal      = {{ACM} Trans. Graph.},
  volume       = {41},
  number       = {4},
  pages        = {141:1--141:13},
  year         = {2022},
  url          = {https://doi.org/10.1145/3528223.3530164},
  doi          = {10.1145/3528223.3530164},
  bibsource    = {dblp computer science bibliography, https://dblp.org}
}

@inproceedings{guo2025shortftdiffusionmodelalignment,
  author       = {Xiefan Guo and
                  Miaomiao Cui and
                  Liefeng Bo and
                  Di Huang},
  title        = {ShortFT: Diffusion Model Alignment via Shortcut-Based Fine-Tuning},
  booktitle    = {{IEEE/CVF} International Conference on Computer Vision, {ICCV} 2025,
                  Honolulu, HI, USA, October 19-25, 2025},
  pages        = {678--687},
  publisher    = {{IEEE}},
  year         = {2025},
  url          = {https://doi.org/10.1109/ICCV51701.2025.00071},
  doi          = {10.1109/ICCV51701.2025.00071},
  bibsource    = {dblp computer science bibliography, https://dblp.org}
}

@inproceedings{clark2024directlyfinetuningdiffusionmodels,
  author       = {Kevin Clark and
                  Paul Vicol and
                  Kevin Swersky and
                  David J. Fleet},
  title        = {Directly Fine-Tuning Diffusion Models on Differentiable Rewards},
  booktitle    = {The Twelfth International Conference on Learning Representations,
                  {ICLR} 2024, Vienna, Austria, May 7-11, 2024},
  publisher    = {OpenReview.net},
  year         = {2024},
  url          = {https://openreview.net/forum?id=1vmSEVL19f},
  bibsource    = {dblp computer science bibliography, https://dblp.org}
}

@inproceedings{xu2023imagerewardlearningevaluatinghuman,
  author       = {Jiazheng Xu and
                  Xiao Liu and
                  Yuchen Wu and
                  Yuxuan Tong and
                  Qinkai Li and
                  Ming Ding and
                  Jie Tang and
                  Yuxiao Dong},
  editor       = {Alice Oh and
                  Tristan Naumann and
                  Amir Globerson and
                  Kate Saenko and
                  Moritz Hardt and
                  Sergey Levine},
  title        = {ImageReward: Learning and Evaluating Human Preferences for Text-to-Image
                  Generation},
  booktitle    = {Advances in Neural Information Processing Systems 36: Annual Conference
                  on Neural Information Processing Systems 2023, NeurIPS 2023, New Orleans,
                  LA, USA, December 10 - 16, 2023},
  year         = {2023},
  url          = {http://papers.nips.cc/paper\_files/paper/2023/hash/33646ef0ed554145eab65f6250fab0c9-Abstract-Conference.html},
  bibsource    = {dblp computer science bibliography, https://dblp.org}
}

@inproceedings{wu2024turboeditinstanttextbasedimage,
  author       = {Zongze Wu and
                  Nicholas I. Kolkin and
                  Jonathan Brandt and
                  Richard Zhang and
                  Eli Shechtman},
  editor       = {Ales Leonardis and
                  Elisa Ricci and
                  Stefan Roth and
                  Olga Russakovsky and
                  Torsten Sattler and
                  G{\"{u}}l Varol},
  title        = {TurboEdit: Instant Text-Based Image Editing},
  booktitle    = {Computer Vision - {ECCV} 2024 - 18th European Conference, Milan, Italy,
                  September 29-October 4, 2024, Proceedings, Part {LXXX}},
  series       = {Lecture Notes in Computer Science},
  volume       = {15138},
  pages        = {365--381},
  publisher    = {Springer},
  year         = {2024},
  url          = {https://doi.org/10.1007/978-3-031-72989-8\_21},
  doi          = {10.1007/978-3-031-72989-8\_21},
  bibsource    = {dblp computer science bibliography, https://dblp.org}
}

@inproceedings{instantedit,
  author       = {Yiming Gong and
                  Zhen Zhu and
                  Minjia Zhang},
  title        = {InstantEdit: Text-Guided Few-Step Image Editing with Piecewise Rectified
                  Flow},
  booktitle    = {{IEEE/CVF} International Conference on Computer Vision, {ICCV} 2025,
                  Honolulu, HI, USA, October 19-25, 2025},
  pages        = {16808--16817},
  publisher    = {{IEEE}},
  year         = {2025},
  url          = {https://doi.org/10.1109/ICCV51701.2025.01561},
  doi          = {10.1109/ICCV51701.2025.01561},
  bibsource    = {dblp computer science bibliography, https://dblp.org}
}

@inproceedings{yin2024onestepdiffusiondistributionmatching,
  author       = {Tianwei Yin and
                  Micha{\"{e}}l Gharbi and
                  Richard Zhang and
                  Eli Shechtman and
                  Fr{\'{e}}do Durand and
                  William T. Freeman and
                  Taesung Park},
  title        = {One-Step Diffusion with Distribution Matching Distillation},
  booktitle    = {{IEEE/CVF} Conference on Computer Vision and Pattern Recognition,
                  {CVPR} 2024, Seattle, WA, USA, June 16-22, 2024},
  pages        = {6613--6623},
  publisher    = {{IEEE}},
  year         = {2024},
  url          = {https://doi.org/10.1109/CVPR52733.2024.00632},
  doi          = {10.1109/CVPR52733.2024.00632},
  bibsource    = {dblp computer science bibliography, https://dblp.org}
}

@inproceedings{samuel2025lightningfastimageinversionediting,
  author       = {Dvir Samuel and
                  Barak Meiri and
                  Haggai Maron and
                  Yoad Tewel and
                  Nir Darshan and
                  Shai Avidan and
                  Gal Chechik and
                  Rami Ben{-}Ari},
  title        = {Lightning-Fast Image Inversion and Editing for Text-to-Image Diffusion
                  Models},
  booktitle    = {The Thirteenth International Conference on Learning Representations,
                  {ICLR} 2025, Singapore, April 24-28, 2025},
  publisher    = {OpenReview.net},
  year         = {2025},
  url          = {https://openreview.net/forum?id=t9l63huPRt},
  bibsource    = {dblp computer science bibliography, https://dblp.org}
}

@inproceedings{tian2025posteditposteriorsamplingefficient,
  author       = {Feng Tian and
                  Yixuan Li and
                  Yichao Yan and
                  Shanyan Guan and
                  Yanhao Ge and
                  Xiaokang Yang},
  title        = {PostEdit: Posterior Sampling for Efficient Zero-Shot Image Editing},
  booktitle    = {The Thirteenth International Conference on Learning Representations,
                  {ICLR} 2025, Singapore, April 24-28, 2025},
  publisher    = {OpenReview.net},
  year         = {2025},
  url          = {https://openreview.net/forum?id=J8YWCBPgx7},
  bibsource    = {dblp computer science bibliography, https://dblp.org}
}

@inproceedings{deutch2024turboedittextbasedimageediting,
  author       = {Gilad Deutch and
                  Rinon Gal and
                  Daniel Garibi and
                  Or Patashnik and
                  Daniel Cohen{-}Or},
  editor       = {Takeo Igarashi and
                  Ariel Shamir and
                  Hao (Richard) Zhang},
  title        = {TurboEdit: Text-Based Image Editing Using Few-Step Diffusion Models},
  booktitle    = {{SIGGRAPH} Asia 2024 Conference Papers, {SA} 2024, Tokyo, Japan, December
                  3-6, 2024},
  pages        = {41:1--41:12},
  publisher    = {{ACM}},
  year         = {2024},
  url          = {https://doi.org/10.1145/3680528.3687612},
  doi          = {10.1145/3680528.3687612},
  bibsource    = {dblp computer science bibliography, https://dblp.org}
}

@inproceedings{starodubcev2024invertibleconsistencydistillationtextguided,
  author       = {Nikita Starodubcev and
                  Mikhail Khoroshikh and
                  Artem Babenko and
                  Dmitry Baranchuk},
  editor       = {Amir Globersons and
                  Lester Mackey and
                  Danielle Belgrave and
                  Angela Fan and
                  Ulrich Paquet and
                  Jakub M. Tomczak and
                  Cheng Zhang},
  title        = {Invertible Consistency Distillation for Text-Guided Image Editing
                  in Around 7 Steps},
  booktitle    = {Advances in Neural Information Processing Systems 37: Annual Conference
                  on Neural Information Processing Systems 2024, NeurIPS 2024, Vancouver,
                  BC, Canada, December 10 - 15, 2024},
  year         = {2024},
  url          = {http://papers.nips.cc/paper\_files/paper/2024/hash/1704fe7aaff33a54802b83a016050ab8-Abstract-Conference.html},
  bibsource    = {dblp computer science bibliography, https://dblp.org}
}

@inproceedings{xu2023inversionfreeimageeditingnatural,
  author       = {Sihan Xu and
                  Yidong Huang and
                  Jiayi Pan and
                  Ziqiao Ma and
                  Joyce Chai},
  title        = {Inversion-Free Image Editing with Language-Guided Diffusion Models},
  booktitle    = {{IEEE/CVF} Conference on Computer Vision and Pattern Recognition,
                  {CVPR} 2024, Seattle, WA, USA, June 16-22, 2024},
  pages        = {9454--9461},
  publisher    = {{IEEE}},
  year         = {2024},
  url          = {https://doi.org/10.1109/CVPR52733.2024.00903},
  doi          = {10.1109/CVPR52733.2024.00903},
  bibsource    = {dblp computer science bibliography, https://dblp.org}
}

@inproceedings{Mokady_2023_CVPR,
  author       = {Ron Mokady and
                  Amir Hertz and
                  Kfir Aberman and
                  Yael Pritch and
                  Daniel Cohen{-}Or},
  title        = {Null-text Inversion for Editing Real Images using Guided Diffusion
                  Models},
  booktitle    = {{IEEE/CVF} Conference on Computer Vision and Pattern Recognition,
                  {CVPR} 2023, Vancouver, BC, Canada, June 17-24, 2023},
  pages        = {6038--6047},
  publisher    = {{IEEE}},
  year         = {2023},
  url          = {https://doi.org/10.1109/CVPR52729.2023.00585},
  doi          = {10.1109/CVPR52729.2023.00585},
  bibsource    = {dblp computer science bibliography, https://dblp.org}
}

@inproceedings{ju2023directinversionboostingdiffusionbased,
  author       = {Xuan Ju and
                  Ailing Zeng and
                  Yuxuan Bian and
                  Shaoteng Liu and
                  Qiang Xu},
  title        = {PnP Inversion: Boosting Diffusion-based Editing with 3 Lines of Code},
  booktitle    = {The Twelfth International Conference on Learning Representations,
                  {ICLR} 2024, Vienna, Austria, May 7-11, 2024},
  publisher    = {OpenReview.net},
  year         = {2024},
  url          = {https://openreview.net/forum?id=FoMZ4ljhVw},
  bibsource    = {dblp computer science bibliography, https://dblp.org}
}

@inproceedings{song2022denoisingdiffusionimplicitmodels,
  author       = {Jiaming Song and
                  Chenlin Meng and
                  Stefano Ermon},
  title        = {Denoising Diffusion Implicit Models},
  booktitle    = {9th International Conference on Learning Representations, {ICLR} 2021,
                  Virtual Event, Austria, May 3-7, 2021},
  publisher    = {OpenReview.net},
  year         = {2021},
  url          = {https://openreview.net/forum?id=St1giarCHLP},
  bibsource    = {dblp computer science bibliography, https://dblp.org}
}

@inproceedings{garibi2024renoiserealimageinversion,
  author       = {Daniel Garibi and
                  Or Patashnik and
                  Andrey Voynov and
                  Hadar Averbuch{-}Elor and
                  Daniel Cohen{-}Or},
  editor       = {Ales Leonardis and
                  Elisa Ricci and
                  Stefan Roth and
                  Olga Russakovsky and
                  Torsten Sattler and
                  G{\"{u}}l Varol},
  title        = {ReNoise: Real Image Inversion Through Iterative Noising},
  booktitle    = {Computer Vision - {ECCV} 2024 - 18th European Conference, Milan, Italy,
                  September 29-October 4, 2024, Proceedings, Part {XIV}},
  series       = {Lecture Notes in Computer Science},
  volume       = {15072},
  pages        = {395--413},
  publisher    = {Springer},
  year         = {2024},
  url          = {https://doi.org/10.1007/978-3-031-72630-9\_23},
  doi          = {10.1007/978-3-031-72630-9\_23},
  bibsource    = {dblp computer science bibliography, https://dblp.org}
}

@inproceedings{lin2015microsoftcococommonobjects,
  author       = {Tsung{-}Yi Lin and
                  Michael Maire and
                  Serge J. Belongie and
                  James Hays and
                  Pietro Perona and
                  Deva Ramanan and
                  Piotr Doll{\'{a}}r and
                  C. Lawrence Zitnick},
  editor       = {David J. Fleet and
                  Tom{\'{a}}s Pajdla and
                  Bernt Schiele and
                  Tinne Tuytelaars},
  title        = {Microsoft {COCO:} Common Objects in Context},
  booktitle    = {Computer Vision - {ECCV} 2014 - 13th European Conference, Zurich,
                  Switzerland, September 6-12, 2014, Proceedings, Part {V}},
  series       = {Lecture Notes in Computer Science},
  volume       = {8693},
  pages        = {740--755},
  publisher    = {Springer},
  year         = {2014},
  url          = {https://doi.org/10.1007/978-3-319-10602-1\_48},
  doi          = {10.1007/978-3-319-10602-1\_48},
  bibsource    = {dblp computer science bibliography, https://dblp.org}
}

@inproceedings{hertz2023deltadenoisingscore,
  author       = {Amir Hertz and
                  Kfir Aberman and
                  Daniel Cohen{-}Or},
  title        = {Delta Denoising Score},
  booktitle    = {{IEEE/CVF} International Conference on Computer Vision, {ICCV} 2023,
                  Paris, France, October 1-6, 2023},
  pages        = {2328--2337},
  publisher    = {{IEEE}},
  year         = {2023},
  url          = {https://doi.org/10.1109/ICCV51070.2023.00221},
  doi          = {10.1109/ICCV51070.2023.00221},
  bibsource    = {dblp computer science bibliography, https://dblp.org}
}

@article{luo2023latentconsistencymodelssynthesizing,
  author       = {Simian Luo and
                  Yiqin Tan and
                  Longbo Huang and
                  Jian Li and
                  Hang Zhao},
  title        = {Latent Consistency Models: Synthesizing High-Resolution Images with
                  Few-Step Inference},
  journal      = {CoRR},
  volume       = {abs/2310.04378},
  year         = {2023},
  url          = {https://doi.org/10.48550/arXiv.2310.04378},
  doi          = {10.48550/ARXIV.2310.04378},
  eprinttype   = {arXiv},
  eprint       = {2310.04378},
  bibsource    = {dblp computer science bibliography, https://dblp.org}
}

@inproceedings{miyake2024negativepromptinversionfastimage,
  author       = {Daiki Miyake and
                  Akihiro Iohara and
                  Yu Saito and
                  Toshiyuki Tanaka},
  title        = {Negative-Prompt Inversion: Fast Image Inversion for Editing with Text-Guided
                  Diffusion Models},
  booktitle    = {{IEEE/CVF} Winter Conference on Applications of Computer Vision, {WACV}
                  2025, Tucson, AZ, USA, February 26 - March 6, 2025},
  pages        = {2063--2072},
  publisher    = {{IEEE}},
  year         = {2025},
  url          = {https://doi.org/10.1109/WACV61041.2025.00207},
  doi          = {10.1109/WACV61041.2025.00207},
  bibsource    = {dblp computer science bibliography, https://dblp.org}
}

@inproceedings{ren2024hypersdtrajectorysegmentedconsistency,
  author       = {Yuxi Ren and
                  Xin Xia and
                  Yanzuo Lu and
                  Jiacheng Zhang and
                  Jie Wu and
                  Pan Xie and
                  Xing Wang and
                  Xuefeng Xiao},
  editor       = {Amir Globersons and
                  Lester Mackey and
                  Danielle Belgrave and
                  Angela Fan and
                  Ulrich Paquet and
                  Jakub M. Tomczak and
                  Cheng Zhang},
  title        = {Hyper-SD: Trajectory Segmented Consistency Model for Efficient Image
                  Synthesis},
  booktitle    = {Advances in Neural Information Processing Systems 37: Annual Conference
                  on Neural Information Processing Systems 2024, NeurIPS 2024, Vancouver,
                  BC, Canada, December 10 - 15, 2024},
  year         = {2024},
  url          = {http://papers.nips.cc/paper\_files/paper/2024/hash/d4e1c24ac41ff0b82ca1b171731f0b23-Abstract-Conference.html},
  bibsource    = {dblp computer science bibliography, https://dblp.org}
}

@inproceedings{ho2020denoisingdiffusionprobabilisticmodels,
  author       = {Jonathan Ho and
                  Ajay Jain and
                  Pieter Abbeel},
  editor       = {Hugo Larochelle and
                  Marc'Aurelio Ranzato and
                  Raia Hadsell and
                  Maria{-}Florina Balcan and
                  Hsuan{-}Tien Lin},
  title        = {Denoising Diffusion Probabilistic Models},
  booktitle    = {Advances in Neural Information Processing Systems 33: Annual Conference
                  on Neural Information Processing Systems 2020, NeurIPS 2020, December
                  6-12, 2020, virtual},
  year         = {2020},
  url          = {https://proceedings.neurips.cc/paper/2020/hash/4c5bcfec8584af0d967f1ab10179ca4b-Abstract.html},
  bibsource    = {dblp computer science bibliography, https://dblp.org}
}

@inproceedings{song2021scorebasedgenerativemodelingstochastic,
  author       = {Yang Song and
                  Jascha Sohl{-}Dickstein and
                  Diederik P. Kingma and
                  Abhishek Kumar and
                  Stefano Ermon and
                  Ben Poole},
  title        = {Score-Based Generative Modeling through Stochastic Differential Equations},
  booktitle    = {9th International Conference on Learning Representations, {ICLR} 2021,
                  Virtual Event, Austria, May 3-7, 2021},
  publisher    = {OpenReview.net},
  year         = {2021},
  url          = {https://openreview.net/forum?id=PxTIG12RRHS},
  bibsource    = {dblp computer science bibliography, https://dblp.org}
}

@inproceedings{brooks2023instructpix2pixlearningfollowimage,
  author       = {Tim Brooks and
                  Aleksander Holynski and
                  Alexei A. Efros},
  title        = {InstructPix2Pix: Learning to Follow Image Editing Instructions},
  booktitle    = {{IEEE/CVF} Conference on Computer Vision and Pattern Recognition,
                  {CVPR} 2023, Vancouver, BC, Canada, June 17-24, 2023},
  pages        = {18392--18402},
  publisher    = {{IEEE}},
  year         = {2023},
  url          = {https://doi.org/10.1109/CVPR52729.2023.01764},
  doi          = {10.1109/CVPR52729.2023.01764},
  bibsource    = {dblp computer science bibliography, https://dblp.org}
}

@article{labs2025flux1kontextflowmatching,
  author       = {Black Forest Labs and
                  Stephen Batifol and
                  Andreas Blattmann and
                  Frederic Boesel and
                  Saksham Consul and
                  Cyril Diagne and
                  Tim Dockhorn and
                  Jack English and
                  Zion English and
                  Patrick Esser and
                  Sumith Kulal and
                  Kyle Lacey and
                  Yam Levi and
                  Cheng Li and
                  Dominik Lorenz and
                  Jonas M{\"{u}}ller and
                  Dustin Podell and
                  Robin Rombach and
                  Harry Saini and
                  Axel Sauer and
                  Luke Smith},
  title        = {{FLUX.1} Kontext: Flow Matching for In-Context Image Generation and
                  Editing in Latent Space},
  journal      = {CoRR},
  volume       = {abs/2506.15742},
  year         = {2025},
  url          = {https://doi.org/10.48550/arXiv.2506.15742},
  doi          = {10.48550/ARXIV.2506.15742},
  eprinttype   = {arXiv},
  eprint       = {2506.15742},
  bibsource    = {dblp computer science bibliography, https://dblp.org}
}

@inproceedings{hui2024hqedithighqualitydatasetinstructionbased,
  author       = {Mude Hui and
                  Siwei Yang and
                  Bingchen Zhao and
                  Yichun Shi and
                  Heng Wang and
                  Peng Wang and
                  Cihang Xie and
                  Yuyin Zhou},
  title        = {HQ-Edit: {A} High-Quality Dataset for Instruction-based Image Editing},
  booktitle    = {The Thirteenth International Conference on Learning Representations,
                  {ICLR} 2025, Singapore, April 24-28, 2025},
  publisher    = {OpenReview.net},
  year         = {2025},
  url          = {https://openreview.net/forum?id=mZptYYttFj},
  bibsource    = {dblp computer science bibliography, https://dblp.org}
}

@inproceedings{wang2025tamingrectifiedflowinversion,
  author       = {Jiangshan Wang and
                  Junfu Pu and
                  Zhongang Qi and
                  Jiayi Guo and
                  Yue Ma and
                  Nisha Huang and
                  Yuxin Chen and
                  Xiu Li and
                  Ying Shan},
  editor       = {Aarti Singh and
                  Maryam Fazel and
                  Daniel Hsu and
                  Simon Lacoste{-}Julien and
                  Felix Berkenkamp and
                  Tegan Maharaj and
                  Kiri Wagstaff and
                  Jerry Zhu},
  title        = {Taming Rectified Flow for Inversion and Editing},
  booktitle    = {Forty-second International Conference on Machine Learning, {ICML}
                  2025, Vancouver, BC, Canada, July 13-19, 2025},
  series       = {Proceedings of Machine Learning Research},
  volume       = {267},
  publisher    = {{PMLR} / OpenReview.net},
  year         = {2025},
  url          = {https://proceedings.mlr.press/v267/wang25ce.html},
  bibsource    = {dblp computer science bibliography, https://dblp.org}
}

@inproceedings{kulikov2025floweditinversionfreetextbasedediting,
  author       = {Vladimir Kulikov and
                  Matan Kleiner and
                  Inbar Huberman{-}Spiegelglas and
                  Tomer Michaeli},
  title        = {Flowedit: Inversion-Free Text-Based Editing Using Pre-Trained Flow
                  Models},
  booktitle    = {{IEEE/CVF} International Conference on Computer Vision, {ICCV} 2025,
                  Honolulu, HI, USA, October 19-25, 2025},
  pages        = {19721--19730},
  publisher    = {{IEEE}},
  year         = {2025},
  url          = {https://doi.org/10.1109/ICCV51701.2025.01834},
  doi          = {10.1109/ICCV51701.2025.01834},
  bibsource    = {dblp computer science bibliography, https://dblp.org}
}

@article{zimage2025,
  title   = {Z-Image: An Efficient Image Generation Foundation Model with Single-Stream Diffusion Transformer},
  author  = {Z-Image Team and Cai, Huanqia and Cao, Sihan and Du, Ruoyi and Gao, Peng and Hoi, Steven and Hou, Zhaohui and Huang, Shijie and Jiang, Dengyang and Jin, Xin and Li, Liangchen and Li, Zhen and Li, Zhong-Yu and Liu, David and Liu, Dongyang and Shi, Junhan and Wu, Qilong and Yu, Feng and Zhang, Chi and Zhang, Shifeng and Zhou, Shilin},
  journal = {arXiv preprint arXiv:2511.22699},
  year    = {2025},
  url     = {https://arxiv.org/abs/2511.22699}
}

@inproceedings{lipman2022flow,
  author       = {Yaron Lipman and
                  Ricky T. Q. Chen and
                  Heli Ben{-}Hamu and
                  Maximilian Nickel and
                  Matthew Le},
  title        = {Flow Matching for Generative Modeling},
  booktitle    = {The Eleventh International Conference on Learning Representations,
                  {ICLR} 2023, Kigali, Rwanda, May 1-5, 2023},
  publisher    = {OpenReview.net},
  year         = {2023},
  url          = {https://openreview.net/forum?id=PqvMRDCJT9t},
  bibsource    = {dblp computer science bibliography, https://dblp.org}
}

@inproceedings{liu2022flow,
  author       = {Xingchao Liu and
                  Chengyue Gong and
                  Qiang Liu},
  title        = {Flow Straight and Fast: Learning to Generate and Transfer Data with
                  Rectified Flow},
  booktitle    = {The Eleventh International Conference on Learning Representations,
                  {ICLR} 2023, Kigali, Rwanda, May 1-5, 2023},
  publisher    = {OpenReview.net},
  year         = {2023},
  url          = {https://openreview.net/forum?id=XVjTT1nw5z},
  bibsource    = {dblp computer science bibliography, https://dblp.org}
}

@inproceedings{esser2024scaling,
  author       = {Patrick Esser and
                  Sumith Kulal and
                  Andreas Blattmann and
                  Rahim Entezari and
                  Jonas M{\"{u}}ller and
                  Harry Saini and
                  Yam Levi and
                  Dominik Lorenz and
                  Axel Sauer and
                  Frederic Boesel and
                  Dustin Podell and
                  Tim Dockhorn and
                  Zion English and
                  Robin Rombach},
  editor       = {Ruslan Salakhutdinov and
                  Zico Kolter and
                  Katherine A. Heller and
                  Adrian Weller and
                  Nuria Oliver and
                  Jonathan Scarlett and
                  Felix Berkenkamp},
  title        = {Scaling Rectified Flow Transformers for High-Resolution Image Synthesis},
  booktitle    = {Forty-first International Conference on Machine Learning, {ICML} 2024,
                  Vienna, Austria, July 21-27, 2024},
  series       = {Proceedings of Machine Learning Research},
  volume       = {235},
  pages        = {12606--12633},
  publisher    = {{PMLR} / OpenReview.net},
  year         = {2024},
  url          = {https://proceedings.mlr.press/v235/esser24a.html},
  bibsource    = {dblp computer science bibliography, https://dblp.org}
}

@inproceedings{deng2025fireflow,
  author       = {Yingying Deng and
                  Xiangyu He and
                  Changwang Mei and
                  Peisong Wang and
                  Fan Tang},
  editor       = {Aarti Singh and
                  Maryam Fazel and
                  Daniel Hsu and
                  Simon Lacoste{-}Julien and
                  Felix Berkenkamp and
                  Tegan Maharaj and
                  Kiri Wagstaff and
                  Jerry Zhu},
  title        = {FireFlow: Fast Inversion of Rectified Flow for Image Semantic Editing},
  booktitle    = {Forty-second International Conference on Machine Learning, {ICML}
                  2025, Vancouver, BC, Canada, July 13-19, 2025},
  series       = {Proceedings of Machine Learning Research},
  volume       = {267},
  publisher    = {{PMLR} / OpenReview.net},
  year         = {2025},
  url          = {https://proceedings.mlr.press/v267/deng25c.html},
  bibsource    = {dblp computer science bibliography, https://dblp.org}
}

@inproceedings{Nguyen_2025_CVPR,
  author       = {Trong{-}Tung Nguyen and
                  Quang Nguyen and
                  Khoi Nguyen and
                  Anh Tuan Tran and
                  Cuong Pham},
  title        = {SwiftEdit: Lightning Fast Text-Guided Image Editing via One-Step Diffusion},
  booktitle    = {{IEEE/CVF} Conference on Computer Vision and Pattern Recognition,
                  {CVPR} 2025, Nashville, TN, USA, June 11-15, 2025},
  pages        = {21492--21501},
  publisher    = {Computer Vision Foundation / {IEEE}},
  year         = {2025},
  url          = {https://openaccess.thecvf.com/content/CVPR2025/html/Nguyen\_SwiftEdit\_Lightning\_Fast\_Text-Guided\_Image\_Editing\_via\_One-Step\_Diffusion\_CVPR\_2025\_paper.html},
  doi          = {10.1109/CVPR52734.2025.02002},
  bibsource    = {dblp computer science bibliography, https://dblp.org}
}

@misc{jiao2025unieditflowunleashinginversionediting,
    title={UniEdit-Flow: Unleashing Inversion and Editing in the Era of Flow Models}, 
    author={Guanlong Jiao and Biqing Huang and Kuan-Chieh Wang and Renjie Liao},
    year={2025},
    eprint={2504.13109},
    archivePrefix={arXiv},
    primaryClass={cs.CV},
    url={https://arxiv.org/abs/2504.13109}, 
}

@inproceedings{dao2024swiftbrushv2,
  author       = {Trung Tuan Dao and
                  Thuan Hoang Nguyen and
                  Thanh Le and
                  Duc Hong Vu and
                  Khoi Nguyen and
                  Cuong Pham and
                  Anh Tuan Tran},
  editor       = {Ales Leonardis and
                  Elisa Ricci and
                  Stefan Roth and
                  Olga Russakovsky and
                  Torsten Sattler and
                  G{\"{u}}l Varol},
  title        = {SwiftBrush {V2:} Make Your One-Step Diffusion Model Better Than Its
                  Teacher},
  booktitle    = {Computer Vision - {ECCV} 2024 - 18th European Conference, Milan, Italy,
                  September 29-October 4, 2024, Proceedings, Part {LXXXII}},
  series       = {Lecture Notes in Computer Science},
  volume       = {15140},
  pages        = {176--192},
  publisher    = {Springer},
  year         = {2024},
  url          = {https://doi.org/10.1007/978-3-031-73007-8\_11},
  doi          = {10.1007/978-3-031-73007-8\_11},
  bibsource    = {dblp computer science bibliography, https://dblp.org}
}

@article{lu2026chordedit,
  title={ChordEdit: One-Step Low-Energy Transport for Image Editing},
  author={Lu, Liangsi and Chen, Xuhang and Guo, Minzhe and Li, Shichu and Wang, Jingchao and Shi, Yang},
  journal={arXiv preprint arXiv:2602.19083},
  year={2026}
}
\clearpage
\setcounter{page}{1}
\appendix
\begin{center}
    {\Large Appendix}
\end{center}

\section{Fine-tune setup}
\label{sec:fine-tune_setup}
All backbone models were initialized from publicly available checkpoints. We used the
default inference configuration for each accelerated model, namely 4 sampling steps for
SD1.5 and SDXL-Turbo, and 8 sampling steps for Z-Image Turbo. For each backbone, we
fine-tuned two LoRA adapters of identical architecture, corresponding to the forward
model $F_\phi$ and the backward model $G_\theta$. The LoRA rank was set to $32$ for
SD1.5 and $64$ for SDXL-Turbo and Z-Image Turbo, with adapters applied to the attention
projections \texttt{[to\_k, to\_v, to\_q, to\_out.0]}. All loss terms used a weight of
$1.0$, except the KL term, which was weighted by $0.5$ for SD1.5 and $0.1$ for
SDXL-Turbo and Z-Image Turbo; the low-frequency term $\mathcal{L}_{\mathrm{LF}}$ used a
pooling size $k=8$. Classifier-free guidance was disabled during fine-tuning. The overall
scheme of our pipeline is shown in Figure~\ref{fig:framework}. All models
were optimized with AdamW at a learning rate of $4\times10^{-4}$ under a cosine schedule,
on 4 NVIDIA H100 GPUs. We used a batch size of $16$ for SD1.5 and SDXL-Turbo, and $4$ for
Z-Image Turbo. Fine-tuning was run until convergence, which took $300$ iterations for SD1.5 and $600$
for SDXL-Turbo and Z-Image Turbo; we use the final checkpoint directly. The full recipe costs $\approx$ 15 H100 GPU-hours once per backbone (300 iterations for SD1.5, 600 for SDXL-Turbo and Z-Image Turbo).

\textbf{Loss-weight sweeps.} Table~\ref{tab:loss_weight_sweeps}
retrains the model at each weight setting and evaluates all runs at
the Table~\ref{tab:main_results} operating point. No setting improves on the default:
every non-default weight lands on or strictly inside the
quality--preservation frontier that the default model already traces
by sweeping $\omega$. Raising $\lambda_{\mathrm{LF}}$ strengthens
edits at a preservation cost, but no better than raising $\omega$
instead ($\lambda_{\mathrm{LF}}{=}2$: dCLIP $0.171$ at LPIPS $0.264$,
vs.\ the default at $\omega{\approx}3.4$: $0.171$ at $0.224$); the
model trains stably across the full $8\times$ range. The KL weight
has a narrower useful band: $\lambda_{\mathrm{KL}}{=}0$ weakens edits
(dCLIP $0.154 \to 0.135$), $\lambda_{\mathrm{KL}}{=}0.05$ is strictly
dominated by the default at $\omega{=}3$ (same dCLIP, LPIPS $0.240$
vs.\ $0.212$), and $\lambda_{\mathrm{KL}}{=}0.5$ collapses training on this backbone
(IR $-2.27$): the prior overwhelms reconstruction and $z_T$ stops
carrying the source. The useful KL band is backbone-dependent: the sweep above is on
Z-Image Turbo, whereas the SD1.5 recipe trains stably at its default
$\lambda_{\mathrm{KL}}{=}0.5$. The per-backbone defaults
are $(\lambda_{\mathrm{LF}}, \lambda_{\mathrm{KL}}) = (1, 0.5)$ for
SD1.5 and $(1, 0.1)$ for SDXL-Turbo and Z-Image Turbo, used
everywhere else in the paper.

\paragraph{Parameter cost and comparison to fine-tuning-based editors.}
Our method trains only two LoRA adapters (forward inversion, backward denoising) on a
frozen backbone. Counting directly from the checkpoints, this is $12.75$M, $185.79$M, and
$133.69$M trainable parameters for SD1.5, SDXL-Turbo, and Z-Image Turbo respectively (Table~\ref{tab:params}),
split evenly across the two adapters and amounting to 1.5--7\% of the
frozen base. This is markedly lighter than prior fine-tuning-based fast editors in
procedure, not just count: iCD~\cite{starodubcev2024invertibleconsistencydistillationtextguided}
obtains its inversion through consistency distillation, and
SwiftEdit~\cite{Nguyen_2025_CVPR} fully trains a dedicated $\sim\!921$M one-step inversion
network in two stages on top of an $\sim\!866$M generator. We add no separate inversion
network and no distillation: the same single-stage recipe (300--600 iterations) applies
across U-Net and DiT backbones.
\subsection{Relation to SwiftEdit}
\label{app:swiftedit}

Our pipeline shares its outer shape with SwiftEdit~\cite{Nguyen_2025_CVPR}:
invert to a latent, regenerate under the target prompt, regularize the
inverted latent. It differs in every component that shape is built from.

\paragraph{What is trained.} SwiftEdit trains a dedicated one-step
inversion network of the same scale as its ${\sim}866$M generator, in
two stages (synthetic images from SwiftBrushv2, then real ones), with
an IP-Adapter-style conditioning branch. We train no new network: two
LoRA adapters ($12.75$--$185.79$M, $1.5$--$7\%$ of the frozen base) on
a single backbone, in one stage of 300--600 iterations on fewer than
10K triplets.

\paragraph{One-sided vs.\ two-sided.} SwiftEdit trains only the
inversion side; the generator enters the loop as a fixed one-step
decoder, and editing is steered at inference by a self-guided mask and
attention rescaling. We adapt both directions and calibrate them to
each other, and our inference uses no attention control and no masks.
Figure~\ref{fig:adapter_ablation} isolates this: training the forward
adapter alone is unstable and settles at lower dCLIP and ImageReward,
because the fixed denoiser is not adapted to the learned terminal
latent.

\paragraph{What the latent regularizer constrains.} Both methods
regularize the inverted latent toward the noise prior, but
channel-space statistics alone are permutation-blind: a standardized
image latent attains channel-KL of $0.000$ while carrying
$16.9\times$ the white-noise low-frequency energy
(Fig.~\ref{fig:klblind}). Our low-frequency hinge
(Eq.~\ref{eq:lf_loss}) closes this blind spot, and
Fig.~\ref{fig:lfcontam} shows the statistic it penalizes is causally
harmful: injecting low-frequency structure at fixed channel-KL
degrades editing monotonically.

\paragraph{Scope.} SwiftEdit is re-instantiable on other one-step
generators (DMD2, InstaFlow, SwiftBrush), but each re-initializes and
retrains the inversion network through both stages, and all
demonstrated backbones remain U-Net (SD1.5/SD2.1). The same
single-stage recipe here transfers unchanged across three distillation
paradigms and to a DiT flow backbone.

\section{Human preference study}
\label{app:human}
\textbf{Protocol.} We use the first $50$ images of PIE-Bench~\citep{ju2023directinversionboostingdiffusionbased} and, for each, edit
it with our method and with every baseline (RF-Edit~\citep{wang2025tamingrectifiedflowinversion}, FlowEdit~\citep{kulikov2025floweditinversionfreetextbasedediting},
FireFlow~\citep{deng2025fireflow}, UniEdit-Flow~\citep{jiao2025unieditflowunleashinginversionediting}). A rater sees the source image and prompt,
the edit prompt, and the two edited results (ours and a baseline) in randomized order, and
answers which they prefer overall, which better matches the edit prompt, and which better
preserves the source; a tie is allowed. This yields $50$ comparisons per baseline ($200$
total), scored as $\mathrm{win}+\tfrac{1}{2}\mathrm{tie}$ ($50\%$ = parity). Each comparison is judged independently by two annotators, who answer all three questions (overall preference, prompt match, source preservation). For any question on which their answers differ, including cases where one chooses a side and the other a tie, an expert rater inspects that comparison and provides the rating, which is taken as final. In total four people participated: two fixed annotators across all 200 comparisons and two expert adjudicators. Prior to adjudication, the two annotators agreed on 97.5\% of question-level judgments.

\textbf{Results.} Our method is preferred over every baseline on overall quality and
preservation, with wins far outnumbering losses ($128/23$ and $138/23$ of $200$). Prompt
match is near parity, dominated by ties, so our edits follow the prompt about as faithfully
as the baselines; the decisive gap is preservation, where baselines reach comparable edit
fidelity only by altering the source. Parity on this question is expected rather than in tension with the
dCLIP gaps in Table~\ref{tab:main_results}: the prompt-match judgment is effectively
binary: raters decide whether the requested edit is present, not how
strongly it is applied, so it saturates once any adequate edit is
made. The automatic metrics show the same compression: under the
official protocol, CLIP$_{\text{ed}}$ spans only $20.7$--$23.3$
across the methods of Table~\ref{tab:main_results}, against a no-edit floor of $20.1$
(scoring the unedited source images), whereas the graded dCLIP
separates the same methods several-fold within each block. This mirrors Table~\ref{tab:human}: we match competing
flow editors on edit strength while preserving more of the input.

\section{Ablations}
\label{app:ablations}
To understand what drives our method, we provide a detailed analysis of each component of
our optimization strategy: the forward and backward adapters (Fig.~\ref{fig:adapter_ablation})
and the individual loss terms (Table~\ref{tab:ablation}). Throughout the appendices, LPIPS$_{\text{wh}}$ (VGG-16) denotes
whole-image VGG-16 LPIPS, and LPIPS-pres/PSNR-pres denote the same
metrics computed outside the edit mask by our internal scorer. Both
differ from the official background-masked SqueezeNet LPIPS and
masked PSNR of Table~\ref{tab:main_results} and are not comparable to them in absolute
value.
\subsection{Single adapter ablation}
\begin{figure}
\centering
\includegraphics[width=\linewidth]{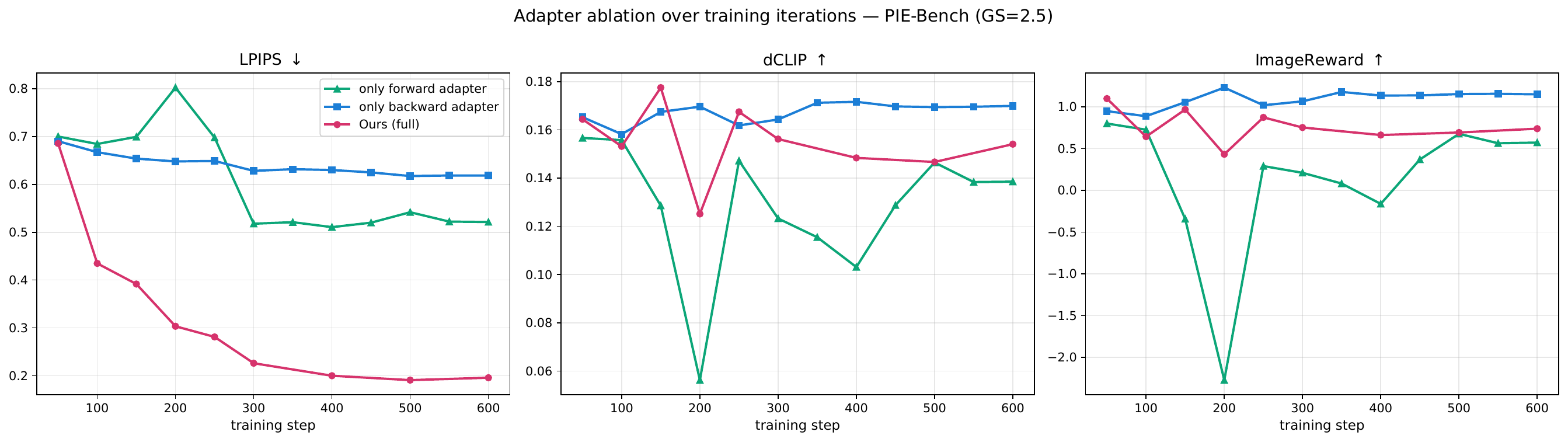}
    \caption{\textbf{Adapter ablation over training iterations} (PIE-Bench, $\omega{=}2.5$).
LPIPS$_{\text{wh}}$(VGG-16)$\downarrow$, dCLIP$\uparrow$, and ImageReward$\uparrow$ for three training
regimes: only the forward adapter $F_\phi$, only the backward adapter $G_\theta$, and both
jointly (full). The backward-only model plateaus at high LPIPS; the forward-only model is
unstable and settles at lower dCLIP and ImageReward. Joint training of both adapters
reaches markedly lower LPIPS while remaining competitive on edit quality, showing that
inversion and generation must be calibrated jointly.}
    \label{fig:adapter_ablation}
\end{figure}

We analyze the contribution of each adapter by fine-tuning only the forward adapter
$F_\phi$, only the backward adapter $G_\theta$, and both jointly (our full method), and
track LPIPS, dCLIP, and ImageReward on PIE-Bench over training
(Fig.~\ref{fig:adapter_ablation}). Neither single-adapter variant is sufficient. Training
only $G_\theta$ leaves reconstruction poor: LPIPS plateaus early and never approaches the
joint model, since the backward pass cannot compensate for an uncalibrated inversion
distribution. Training only $F_\phi$ reaches better LPIPS than the backward-only variant
but is unstable across training and converges to a worse operating point, with lower
CLIP-Edit and ImageReward, because the fixed denoiser is not adapted to the learned
terminal latent. Optimizing both adapters jointly resolves both failure modes, attaining
substantially lower LPIPS while remaining competitive on dCLIP and ImageReward
throughout training, confirming that inversion and generation must be calibrated to each
other rather than tuned in isolation.

\subsection{Analysis of the learned terminal latents}
\label{app:latents}

\begin{figure}[t]
\centering
\includegraphics[width=\linewidth]{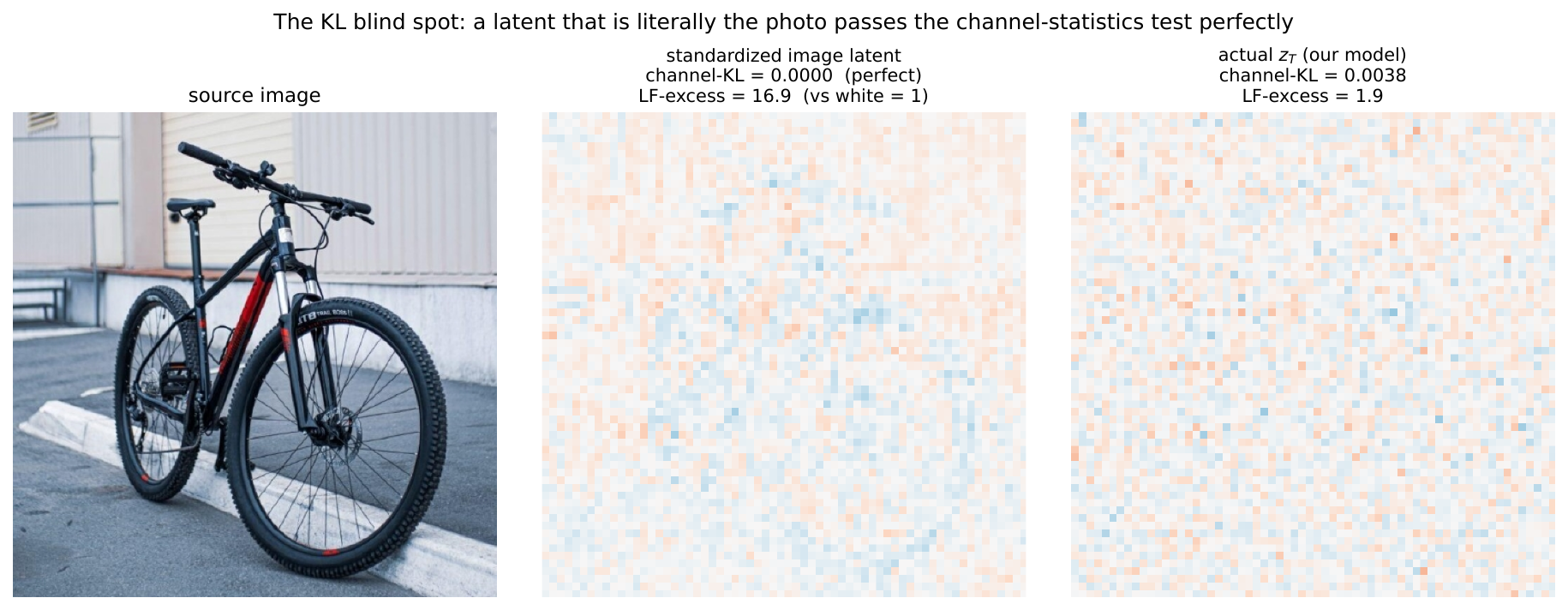}
\caption{\textbf{The KL blind spot.} A per-channel standardized image
latent (middle; the bicycle is faintly visible) passes the
channel-statistics test perfectly (KL $=0.000$) yet carries
$16.9\times$ the white-noise low-frequency energy; our inverted $z_T$
(right) passes both tests.}
\label{fig:klblind}
\end{figure}

\textbf{Complementarity of the two priors.} The priors are
complementary by construction. A per-channel standardized image
latent attains channel-KL of exactly $0$ while its low-frequency
energy is $16.9\times$ the white-noise floor
(Fig.~\ref{fig:klblind}); conversely, under-scaled white noise
($0.77\,\varepsilon$, the empirical channel scale of the base model's
terminal latents) passes the one-sided LF hinge ($0.60$) but fails the
KL term ($0.058$). On MS-COCO ($n{=}2700$) the KL term drives the median per-image
channel-KL of the inverted latents from $0.0186$ (trained without it)
to $0.0032$. Symmetrically for the LF statistic: without the hinge, the pooled
low-frequency excess reaches $4.1\times$ the white-noise floor in the
worst case (median $1.73$); re-inverting that worst case with the
full model removes ${\approx}35\%$ of its low-frequency energy
($4.1\times \to 2.6\times$), and the population median drops to
$1.61$. Increasing $\lambda_{\mathrm{LF}}$ to $2$ suppresses the tail
entirely (99th percentile $2.55 \to 1.85$).

\begin{figure}[t]
\centering
\includegraphics[width=\linewidth]{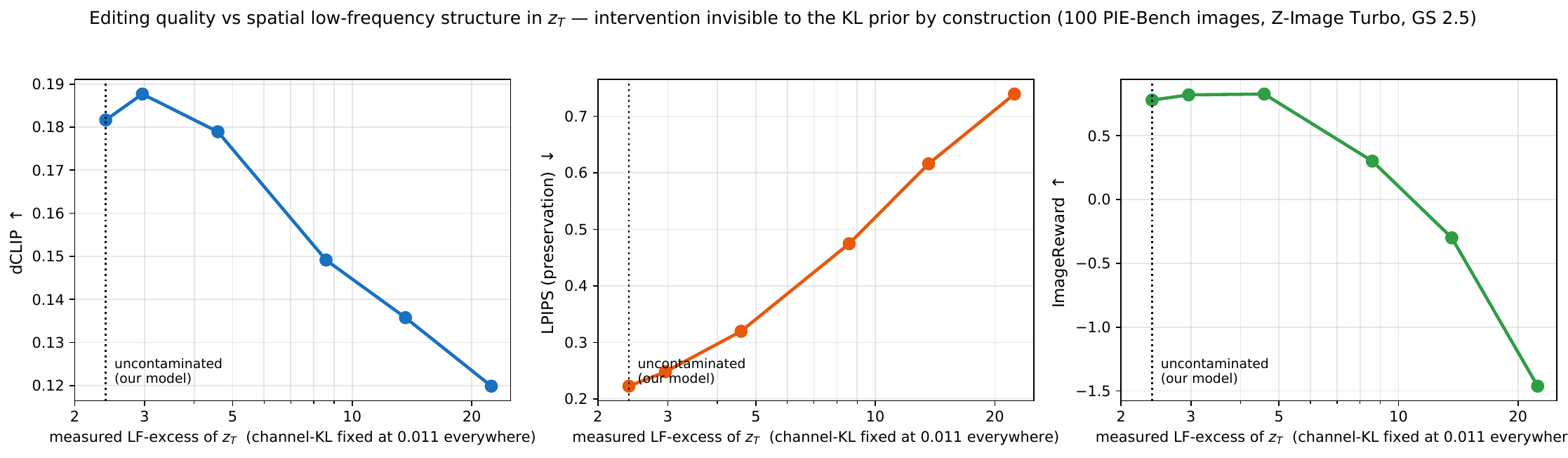}
\caption{\textbf{Causal LF-contamination experiment} (PIE-Bench,
$n{=}100$, $\omega{=}2.5$). Low-frequency noise is injected into
$z_T$ at increasing strength and renormalized per channel, so
channel-KL is unchanged by construction; editing quality nevertheless
degrades monotonically in dCLIP, preservation LPIPS, and ImageReward.
The statistic penalized by the LF hinge is causally harmful, not
merely correlated. Channel-KL here is the mean per-image statistic over the 100
PIE-Bench source images (right-skewed; median $0.006$) and is
unchanged by the intervention by construction; the MS-COCO medians
quoted in Sec.~\ref{app:latents} are computed identically on a different image set.}
\label{fig:lfcontam}
\end{figure}

\textbf{Causal effect of LF structure.} Injecting low-frequency noise
into $z_T$ at increasing strength, renormalized per channel so that
channel-KL is unchanged by construction, monotonically degrades
editing (Fig.~\ref{fig:lfcontam}): the statistic penalized by the LF
hinge is causally harmful, consistent with the frontier ablation
(Fig.~\ref{fig:priors_pareto}), where the without-LF frontier is the
weakest.

\begin{figure}[t]
\centering
\includegraphics[width=0.368\linewidth]{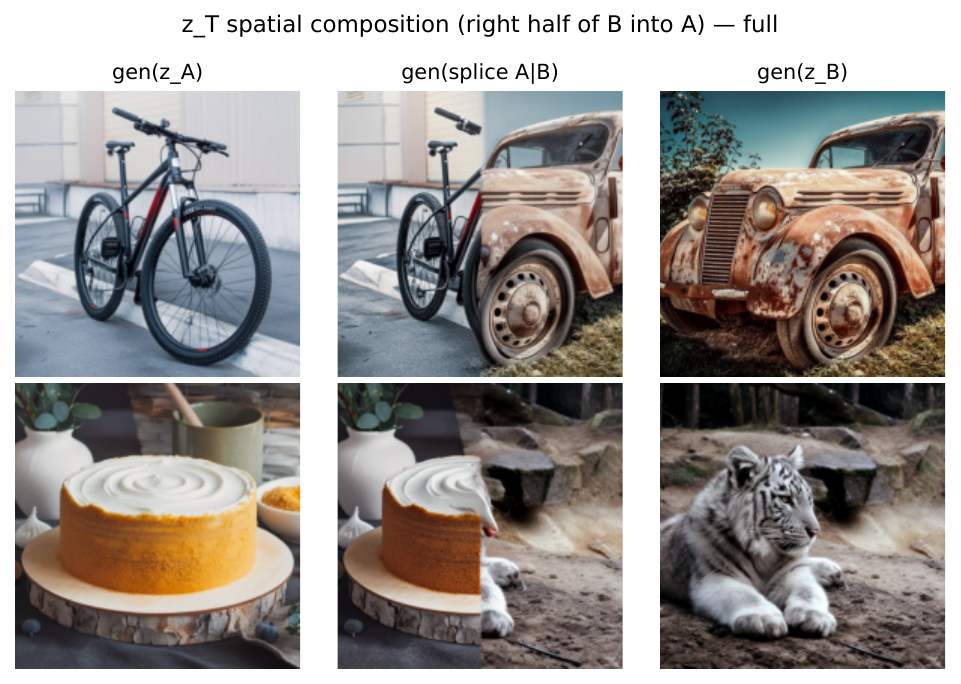}\hfill
\includegraphics[width=0.622\linewidth]{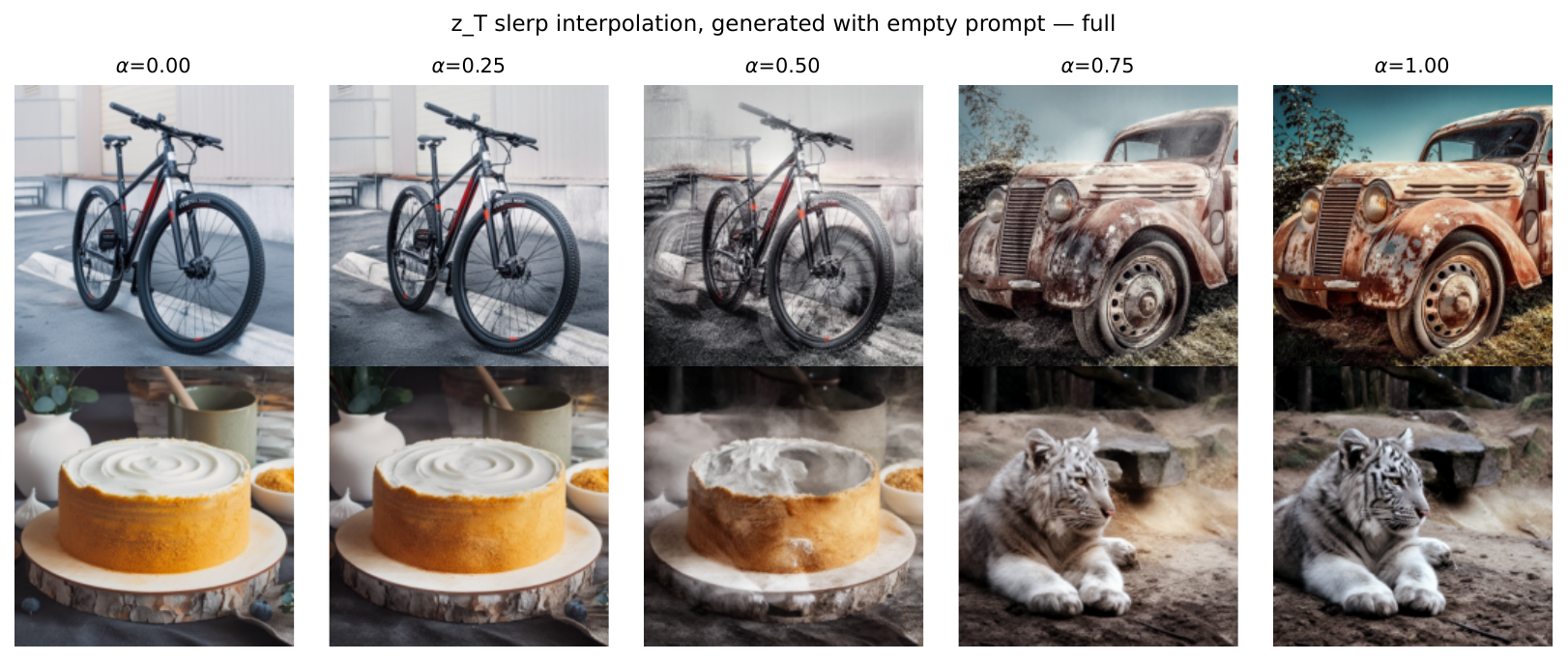}
\caption{\textbf{The learned terminal latents compose and
interpolate.} Left: splicing the right half of $z_B$ into $z_A$
decodes each half near round-trip quality. Right: slerp interpolation
between two inverted latents (decoded with an empty prompt) yields
smooth semantic morphs.}
\label{fig:latentops}
\end{figure}

\textbf{Composition and interpolation.} Splicing two inverted latents
half-and-half (30 PIE-Bench pairs) decodes each half near round-trip
quality: LPIPS $0.195$ vs.\ $0.144$ for full reconstruction of the
same parent images under the same decoding protocol, and $0.775$
against the wrong source. Interpolating pairs of inverted latents
yields monotonic morphs on 97--100\% of pairs with path length
$1.59\times$ the direct distance, vs.\ $2.95\times$ for raw re-noised
latents (Fig.~\ref{fig:latentops}).

\begin{figure}[t]
\centering
\includegraphics[width=\linewidth]{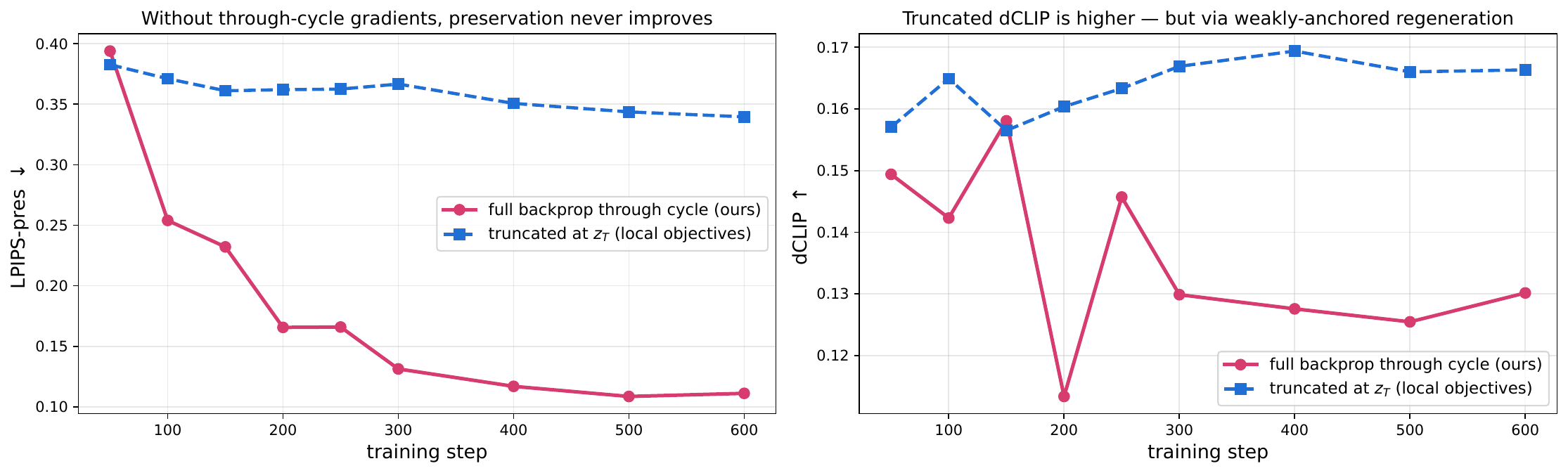}
\caption{\textbf{Gradient-flow ablation:} same recipe with gradients
detached at $z_T$ (Z-Image Turbo, $\omega{=}2.5$). Left: without
through-cycle gradients, preservation never improves. Right: the
truncated model's higher dCLIP reflects weakly-anchored
regeneration from the caption, not editing of the input. LPIPS-pres: VGG-16 LPIPS outside the edit mask.}
\label{fig:truncbp}
\end{figure}

\begin{figure}[t]
\centering
\includegraphics[width=\linewidth]{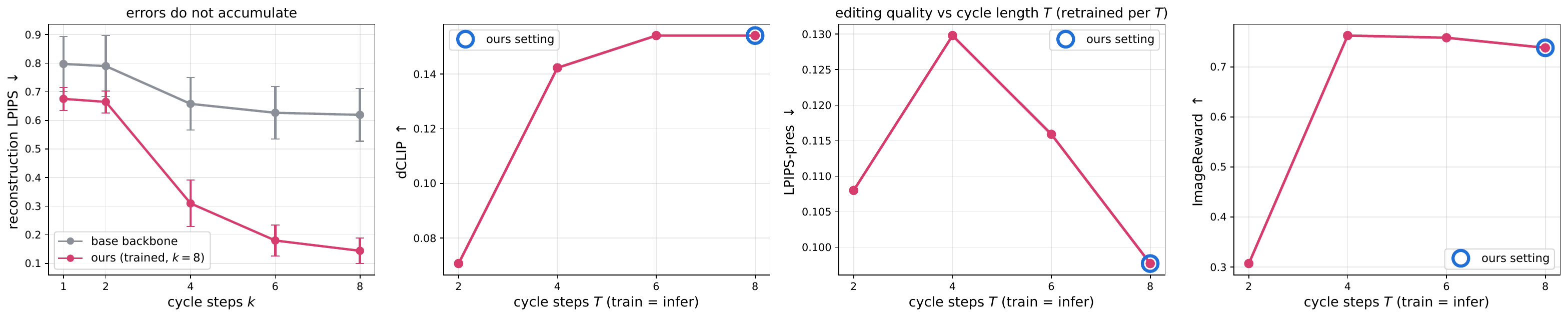}
\caption{\textbf{Cycle length.} Left: reconstruction LPIPS (VGG-16) vs.\ cycle
steps $k$ (PIE-Bench, n=50); errors do not accumulate, the trained model improves with
$k$ while the base backbone stays flat. Right three panels: retraining
the same recipe at $T \in \{2,4,6,8\}$ and evaluating each at its
native $T$; editing saturates at $T{=}6$, preservation peaks at
$T{=}8$ (our setting, circled).}
\label{fig:cyclelen}
\end{figure}

\subsection{Global vs.\ local optimization and cycle length}
\label{app:truncbp}

\textbf{Truncating the gradient at $z_T$.} To isolate the effect of
optimizing through the full cycle, we retrain the Z-Image recipe with
gradients detached at the $z_T$ boundary, so each adapter sees only
its local objective. At our default operating point (guided inference of Eq.~\ref{eq:guided_backward}, $\omega{=}2.5$), the truncated model collapses
reconstruction fidelity: LPIPS-pres 0.338 vs.\ 0.098 for the full
model, PSNR-pres 11.7 vs.\ 24.0. Its dCLIP and ImageReward
nevertheless stay high ($0.174$/$0.77$): with a fixed denoiser
uncalibrated to the learned inversion, the backward pass regenerates
from the caption instead of reconstructing the image. The training
dynamics (Fig.~\ref{fig:truncbp}, shown under uncond CFG) follow the
same pattern: the truncated model's preservation never improves over
600 steps. Global optimization is therefore load-bearing, not merely
convenient.

\textbf{Error does not accumulate with cycle length.} Reconstruction
LPIPS improves, rather than degrades, as the number of cycle steps
grows (Fig.~\ref{fig:cyclelen}, left): the trained model reaches
$0.14$ at $k{=}8$ while the base backbone stays at ${\sim}0.6$ at
every $k$. Retraining the same recipe at $T \in \{2,4,6,8\}$ and
evaluating each at its native $T$ under delta CFG
(Fig.~\ref{fig:cyclelen}, right), editing saturates at $T{=}6$ (dCLIP
$0.154$ at both $T{=}6$ and $8$) while preservation peaks at $T{=}8$
(LPIPS-pres $0.098$ vs.\ $0.116$, PSNR-pres $24.0$ vs.\ $22.9$), our
default.
\subsection{Final objective}
\begin{figure}
\centering
\includegraphics[width=\linewidth]{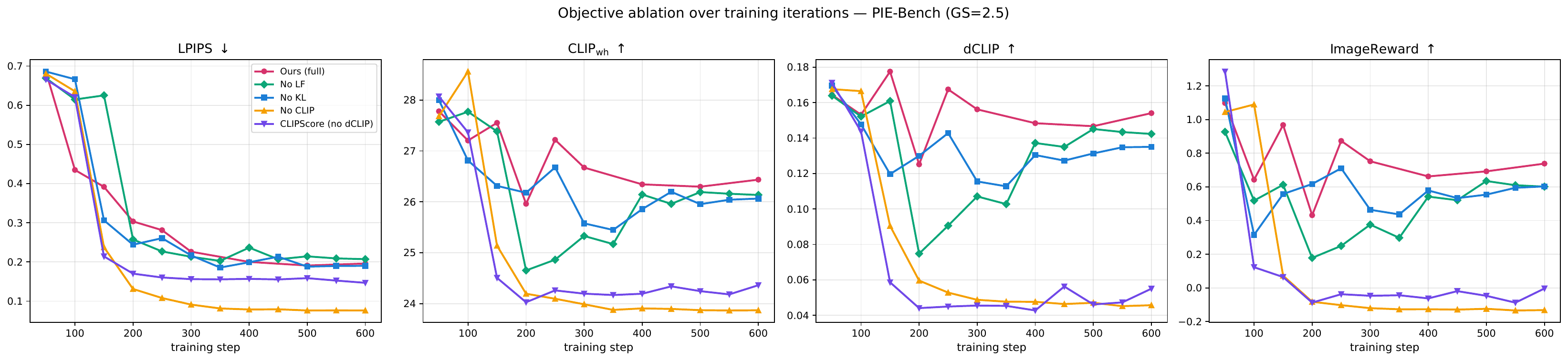}
    \caption{\textbf{Objective ablation on Z-Image Turbo over training iterations}
(PIE-Bench, $\omega{=}2.5$). LPIPS$_{\text{wh}}$(VGG-16)$\downarrow$, CLIP-Score$\uparrow$, dCLIP$\uparrow$,
and ImageReward$\uparrow$ for the full objective and variants that remove the LF term, the
KL term, or the edit term (\textit{No CLIP}), and that replace directional dCLIP with
absolute CLIPScore. Without a directional edit term the model collapses toward the
identity solution (low LPIPS, near-zero dCLIP and ImageReward); the full objective
gives the best edit quality across training.}
    \label{fig:loss_by_iters}
\end{figure}

\begin{figure}
\centering
\includegraphics[width=\linewidth]{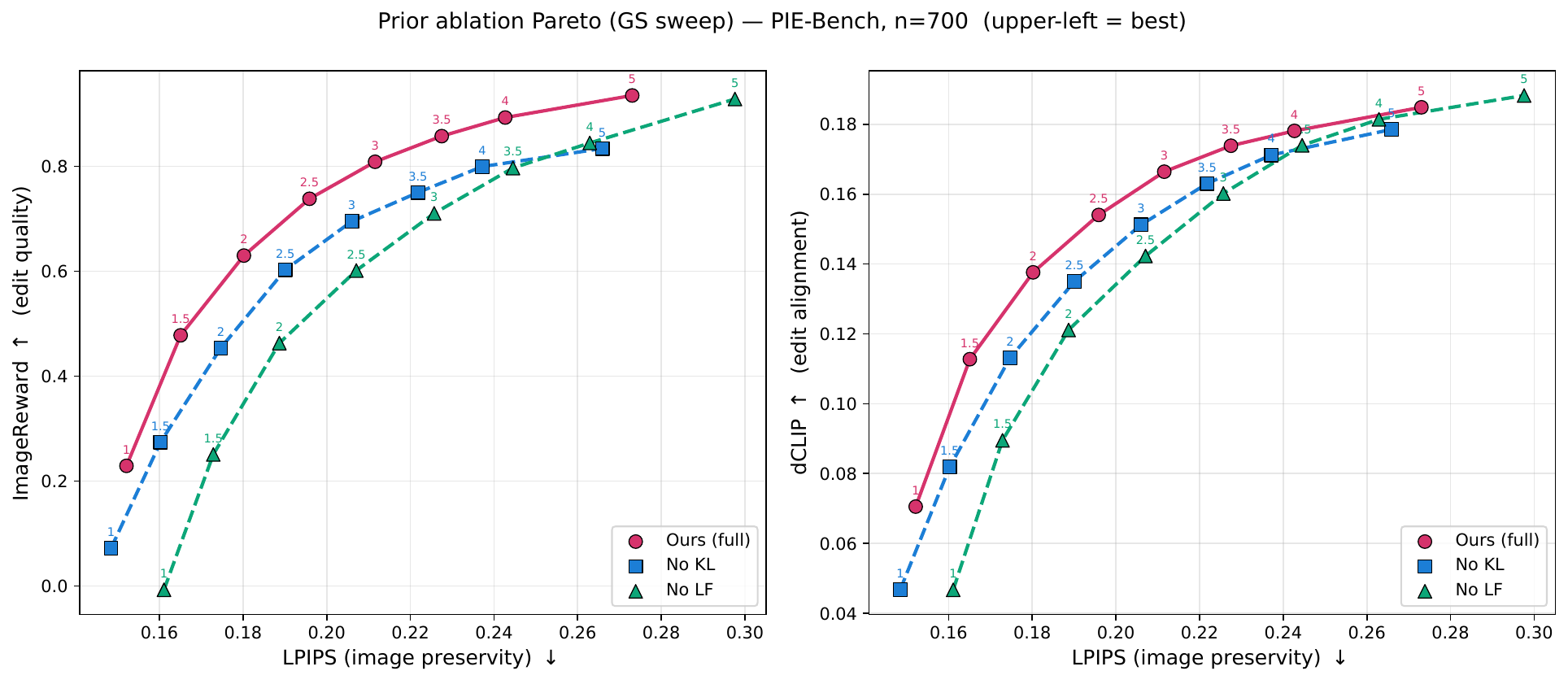}
    \caption{\textbf{Prior-regularizer ablation on Z-Image Turbo: quality--preservation
    frontiers} (PIE-Bench, $n{=}700$, $\omega$ sweep). ImageReward$\uparrow$ (left) and
    dCLIP$\uparrow$ (right) vs.\ LPIPS$_{\text{wh}}$(VGG-16)$\downarrow$; upper-left is best, markers labelled
    with the guidance scale $\omega$. Removing either the KL or the low-frequency (LF) term
    shifts the frontier below and to the right of the full model, with \textit{No LF} the
    weakest.}
    \label{fig:priors_pareto}
\end{figure}

We ablate each term of the training objective on Z-Image Turbo, tracking all metrics on
PIE-Bench over training (Fig.~\ref{fig:loss_by_iters}) together with their effect on
the quality--preservation frontier (Fig.~\ref{fig:priors_pareto}).
\begin{table}[t]
\centering
\small
\caption{Trainable parameters of our two adapters, counted directly from the checkpoints.
The forward and backward LoRA are identical in size; ``\% of base'' is the total trainable
fraction of the frozen backbone.}
\label{tab:params}
\begin{tabular*}{\linewidth}{@{\extracolsep{\fill}}lccccc@{}}
\toprule
Backbone & LoRA rank & Per adapter & Total trainable & Frozen base & \% of base \\
\midrule
SD1.5 (LCM)    & 32 & 6.38\,M  & 12.75\,M  & $\sim$865\,M & $\sim$1.5\% \\
SDXL-Turbo     & 64 & 92.90\,M & 185.79\,M & $\sim$2.6\,B  & $\sim$7\% \\
Z-Image-Turbo  & 64 & 66.85\,M & 133.69\,M & $\sim$6.2B          & $\sim$2.17\% \\
\bottomrule
\end{tabular*}
\end{table}
\emph{Directional CLIP is necessary for editing.} Removing the edit term entirely
(\textit{No CLIP}) drives the model toward the identity solution: it attains the lowest
LPIPS (best preservation) but collapses on dCLIP and ImageReward, reconstructing the
source rather than editing it. Replacing our directional objective with an absolute
CLIP-score objective (\textit{CLIPScore, no dCLIP}) fails in the same way: it produces
little edit signal and tracks close to the no-edit variant on both dCLIP and CLIP
Score, confirming that the edit must be driven by the source-to-target \emph{direction}
rather than absolute alignment.

\emph{The prior regularizers shift the frontier.} Removing either terminal-prior term
degrades the whole quality--preservation trade-off rather than a single operating point
(Fig.~\ref{fig:priors_pareto}): both \textit{No KL} and \textit{No LF} frontiers lie
below and to the right of the full model across the $\omega$ sweep, with \textit{No LF}
the weakest. This indicates that constraining the terminal latent toward the backbone's
noise prior, in both channel statistics (KL) and spatial whiteness (LF), is what lets the
model reach strong edits without sacrificing preservation.
\section{Additional results}
\subsection{Inversion}
\label{app:additional_inversion}
We assess inversion through source reconstruction: we invert an image to the terminal
latent $z_T=f_\phi(x_0,y_\text{src})$ with the forward adapter, then denoise back from
$z_T$ under the same source prompt with the backward adapter, $\hat x_\text{src}=
g_\theta(z_T,y_\text{src})$. No edit is applied, so a faithful method returns the input;
classifier-free guidance is disabled for all methods in this setting. We compare only
against methods that perform an explicit inversion, since inversion-free editors do not
recover a source latent (Table~\ref{tab:reconstruction_results_horizontal}).

\begin{figure}
\centering
\includegraphics[width=\linewidth]{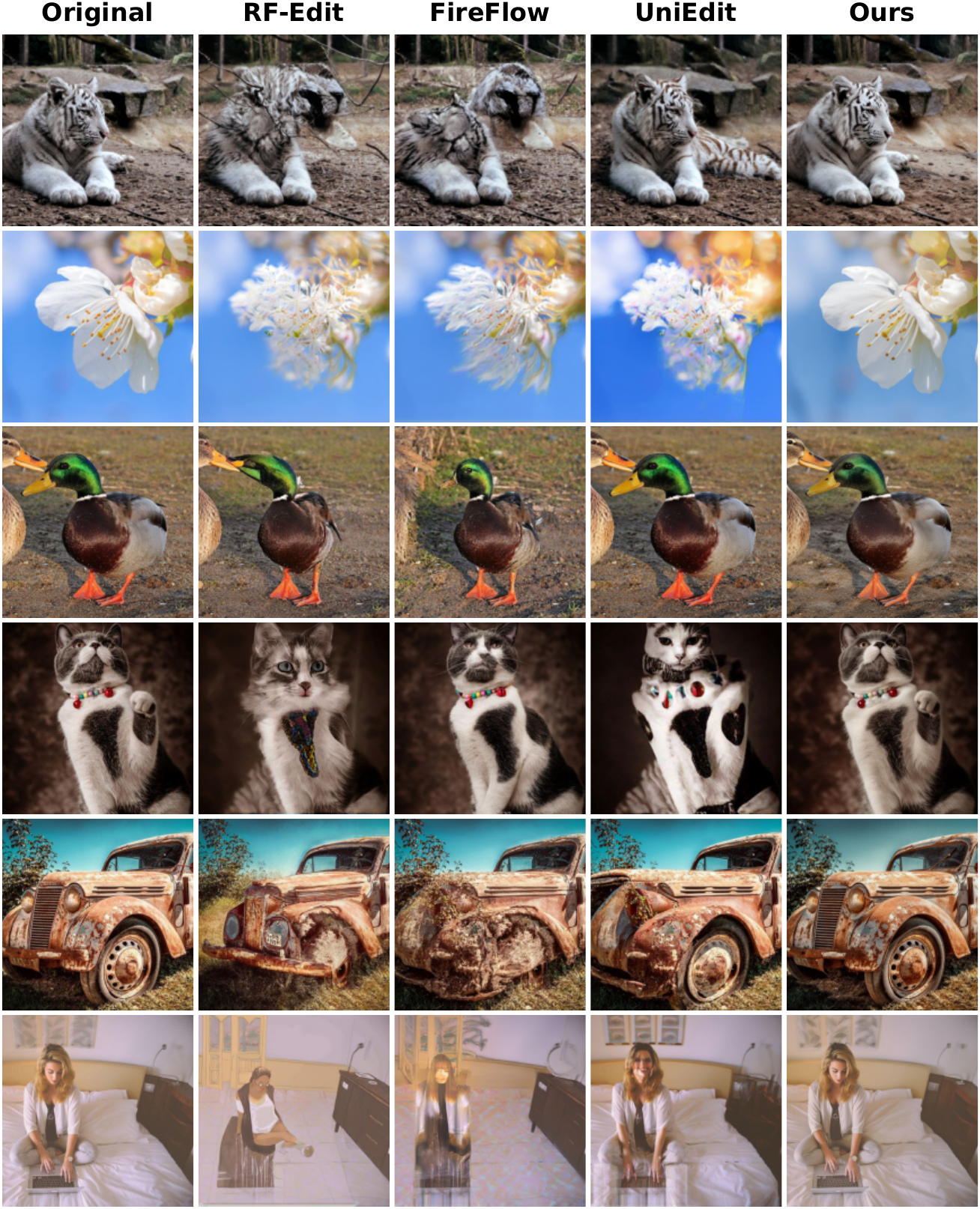}
    \caption{\textbf{Image reconstruction on Z-Image Turbo (flow-based)} against RF-Edit,
FireFlow, and UniEdit-Flow. Each method inverts the source image and reconstructs it under
the source prompt with classifier-free guidance disabled; a faithful method returns the
input. Our adapters recover fine detail, color, and structure most closely.}
    \label{fig:recon_flow}
\end{figure}

\begin{figure}
\centering
\includegraphics[width=0.9\linewidth]{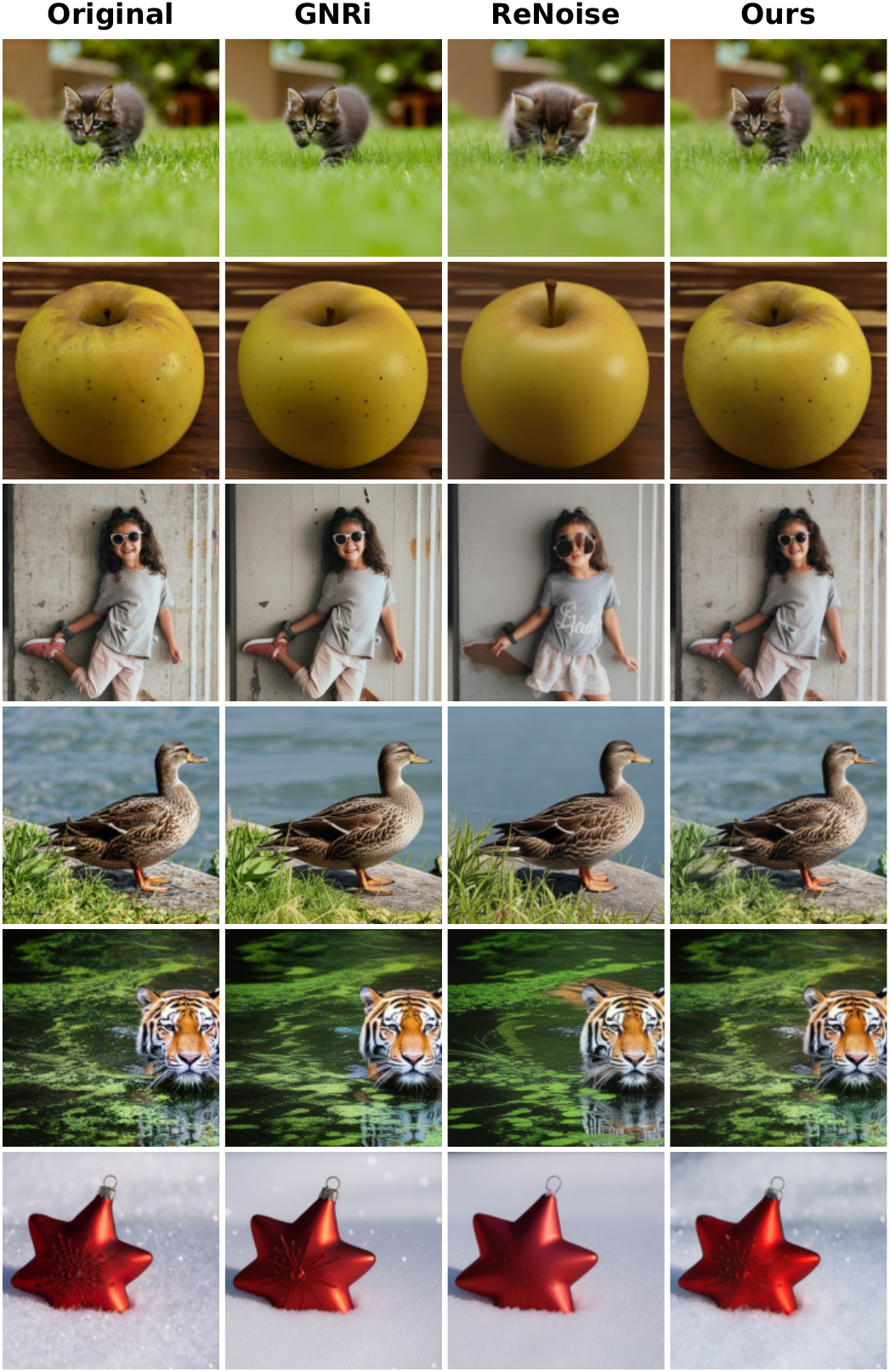}
    \caption{\textbf{Image reconstruction on SDXL-Turbo} against GNRi and ReNoise. Our method
preserves identity and detail, whereas the baselines lose structure or drift in
appearance.}
    \label{fig:recon_sdxl}
\end{figure}

\begin{figure}
\centering
\includegraphics[width=\linewidth]{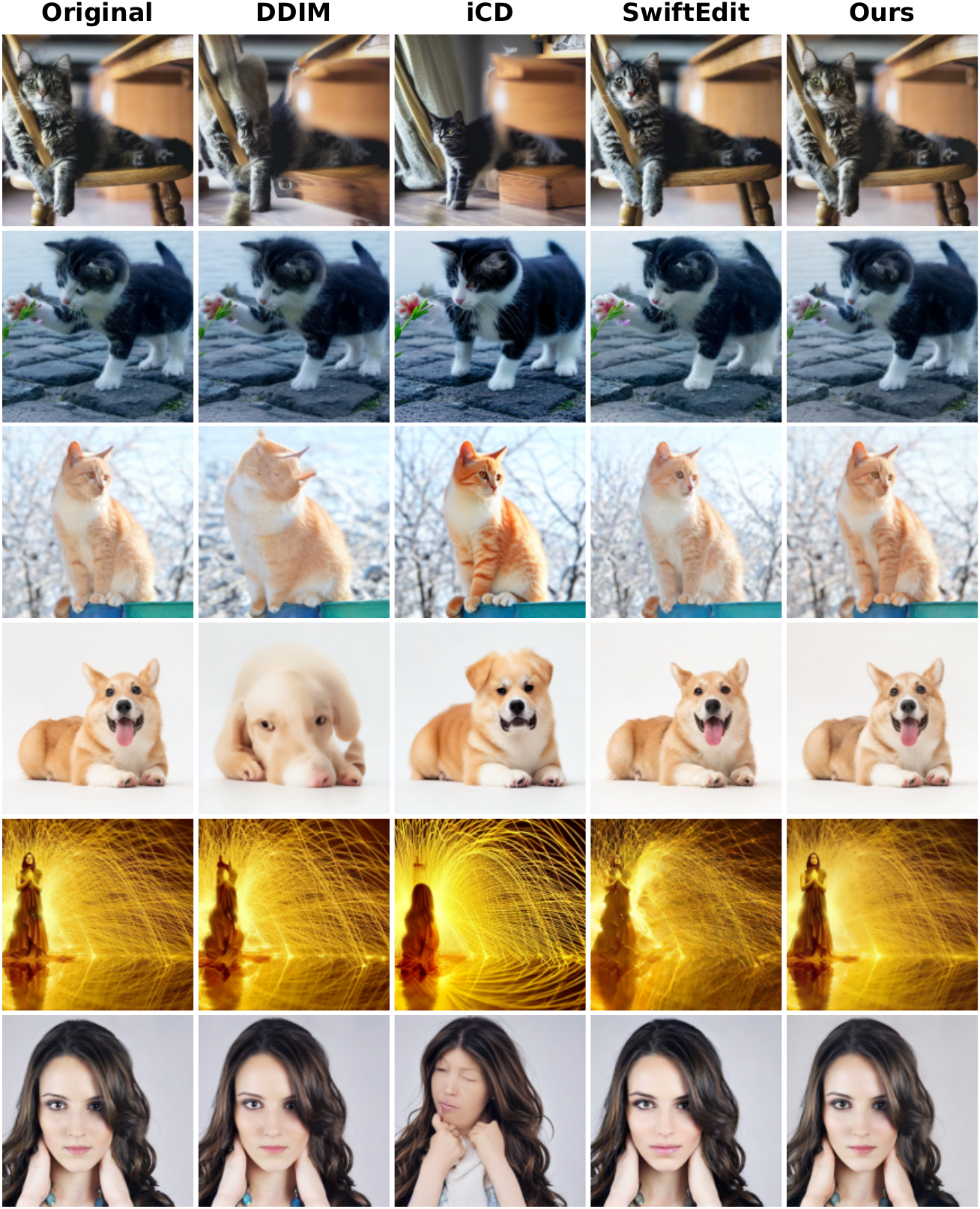}
    \caption{\textbf{Image reconstruction on SD1.5} against full-step DDIM inversion, iCD, and
SwiftEdit. Our method reconstructs the input most faithfully, surpassing even full-step
DDIM despite operating in the few-step regime.}
    \label{fig:recon_sd}
\end{figure}

Quantitatively, our adapters give the most faithful reconstruction in every backbone
family, attaining the lowest MSE and LPIPS, and on the SD backbone surpassing even
full-step DDIM inversion despite operating in the few-step regime
(Table~\ref{tab:reconstruction_results_horizontal}). We show representative
reconstructions per backbone in Figures~\ref{fig:recon_flow}--\ref{fig:recon_sd}:
Z-Image Turbo against the flow-based inversions (Fig.~\ref{fig:recon_flow}), SDXL-Turbo
against the SDXL inversions (Fig.~\ref{fig:recon_sdxl}), and SD1.5 against DDIM, iCD, and
SwiftEdit (Fig.~\ref{fig:recon_sd}). Prior accelerated inversions either lose fine detail
and identity or drift in color and structure, whereas ours preserves the input closely
across diverse content.
\subsection{Editing}
\label{app:additional_editing}
At inference our method uses only the two trained adapters, with no attention control,
masks, or per-image optimization. Given an image $x_0$ with source prompt $y_\text{src}$
and target prompt $y_\text{trg}$, we invert with the forward adapter to obtain
$z_T=f_\phi(x_0,y_\text{src})$, then denoise from the same $z_T$ with the guided backward
adapter $\hat g_\theta(z_T,y_\text{trg},\omega)$ to produce the edit
(Sec.~\ref{sec:editing_inference}); the pipeline is a single
inversion$\rightarrow$generation pass. Both stages are deterministic: no noise is sampled at any
point, so a given $(x_0, y_{\mathrm{src}}, y_{\mathrm{trg}},
\omega)$ always produces the same edit, and results are exactly
reproducible. Consequently the pipeline does not offer diversity by
resampling; controlled variation of the output is obtained by moving
along the $\omega$ axis, which changes edit strength continuously.

\begin{figure}
\centering
\includegraphics[width=\linewidth]{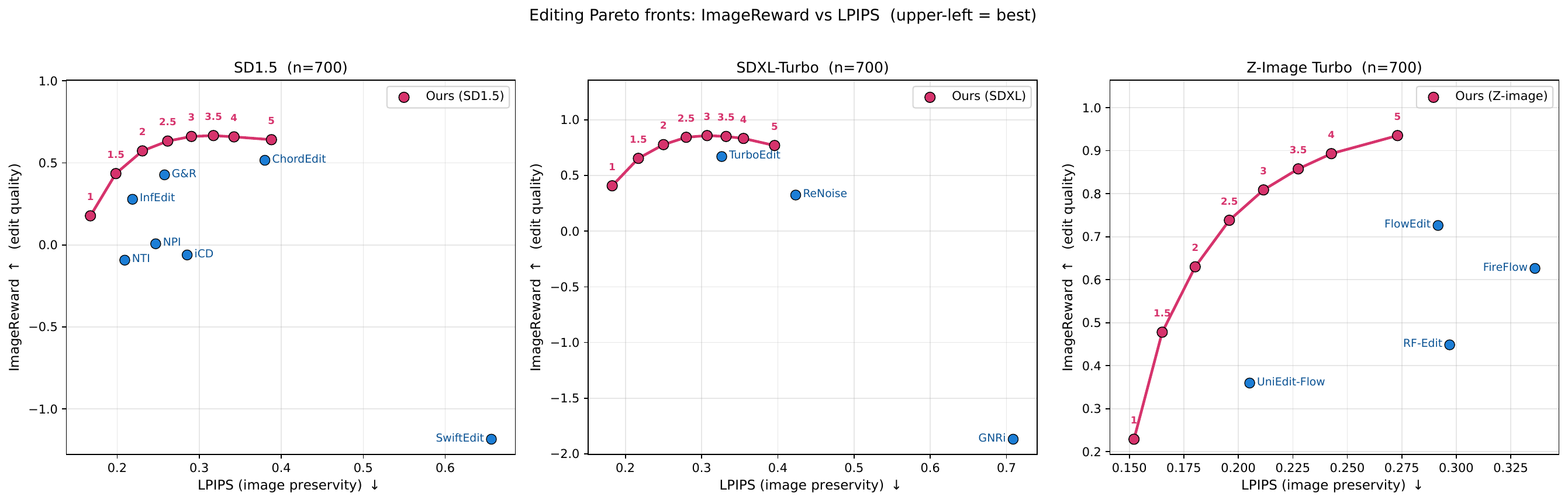}
    \caption{\textbf{Editing quality--preservation frontiers} (PIE-Bench, $n{=}700$):
ImageReward$\uparrow$ vs.\ LPIPS$_{\text{wh}}$(VGG-16)$\downarrow$, upper-left is best. Red curve sweeps our
guidance scale $\omega$ (marker labels); baselines are single points. Our frontier
dominates the baselines on all three backbones (SD1.5, SDXL-Turbo, Z-Image Turbo).}
    \label{fig:pareto_editing}
\end{figure}

Because the edit is controlled by a single guidance scale $\omega$, sweeping $\omega$
traces an entire quality--preservation frontier rather than a single operating point.
Figure~\ref{fig:pareto_editing} plots ImageReward against LPIPS as $\omega$ varies (markers
labelled with $\omega$) on all three backbones. Across every backbone our frontier lies
above and to the left of the baseline operating points: for any preservation level a
baseline reaches, our method attains higher edit quality at equal or lower LPIPS. Since
each baseline is a single fixed point dominated by our curve, the trade-off reported at
our chosen $\omega$ in Table~\ref{tab:main_results} reflects the whole operating range; we
use $\omega=2.0$ for SD1.5, $\omega=3.0$ for SDXL-Turbo, and $\omega=2.5$ for Z-Image
Turbo.

\begin{figure}
\centering
\includegraphics[width=\linewidth]{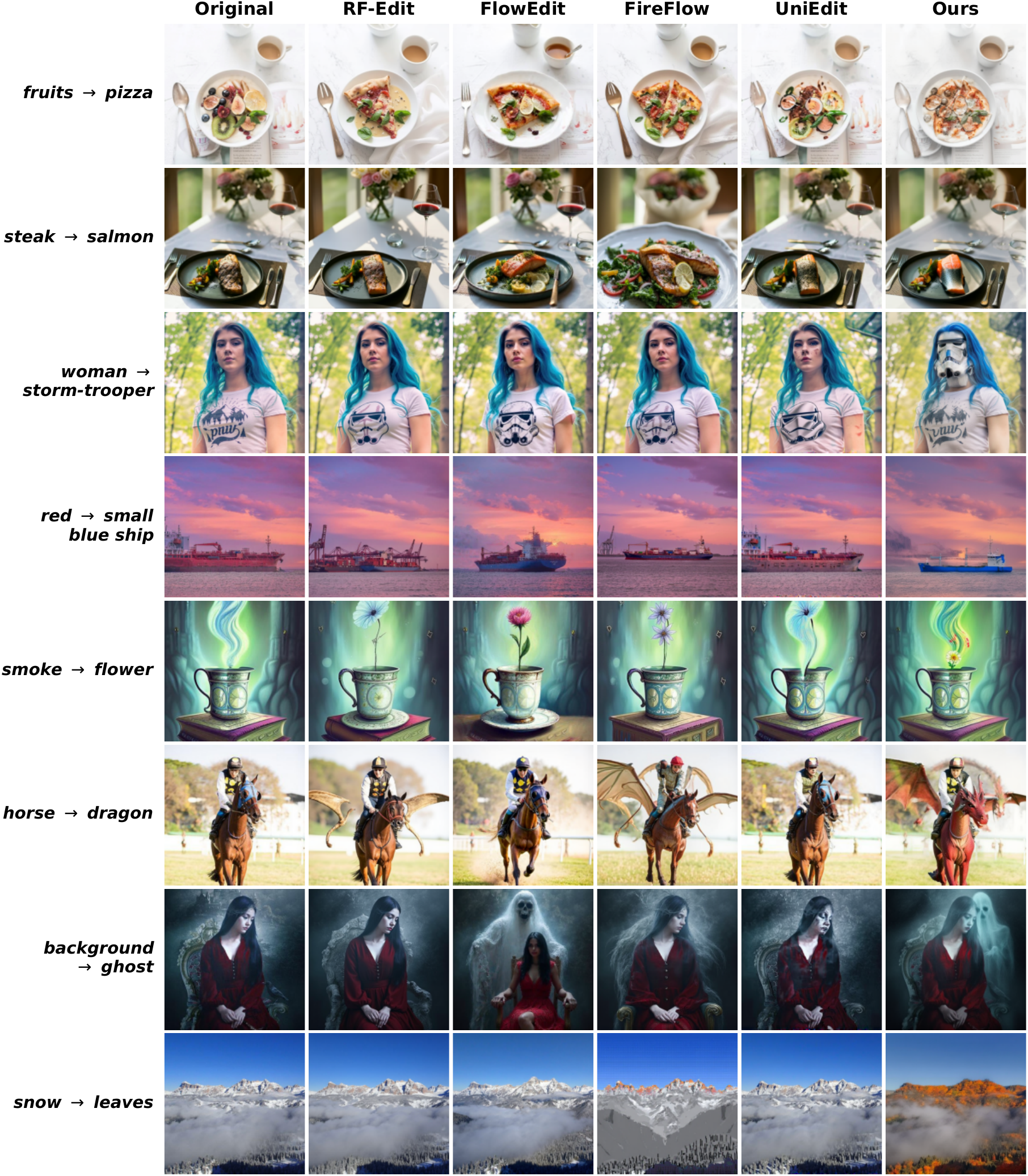}
    \caption{\textbf{Additional qualitative results on Z-Image Turbo (flow-based)} against
RF-Edit, FlowEdit, FireFlow, and UniEdit-Flow (one edit per row). Our method applies each
edit while preserving structure, identity, and background.}
    \label{fig:editing_flow_appendix}
\end{figure}

\begin{figure}
\centering
\includegraphics[width=\linewidth]{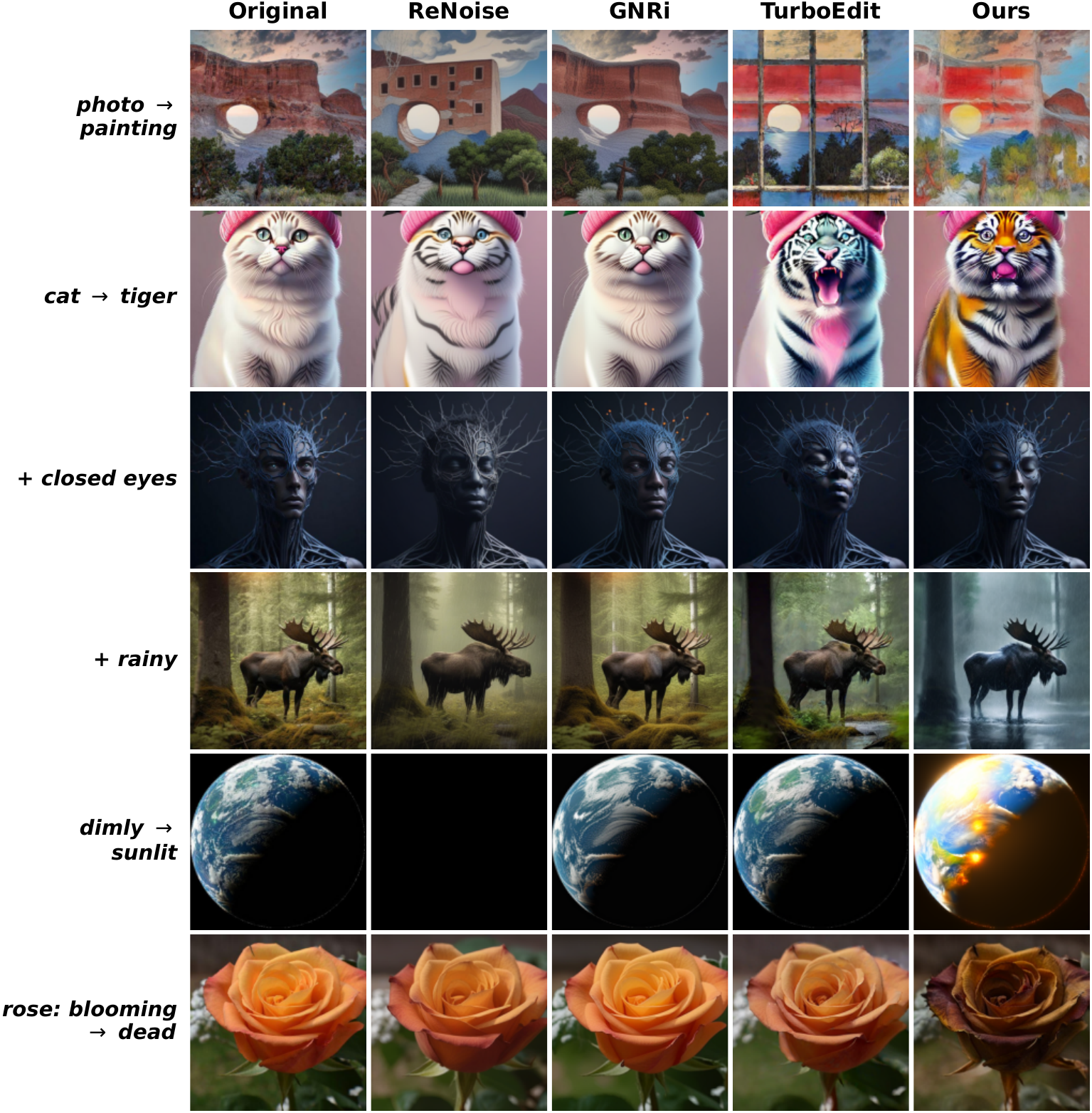}
    \caption{\textbf{Additional qualitative results on SDXL-Turbo} against ReNoise, GNRi, and
TurboEdit (one edit per row).}
    \label{fig:editing_sdxl_appendix}
\end{figure}

\begin{figure}
\centering
\includegraphics[width=\linewidth]{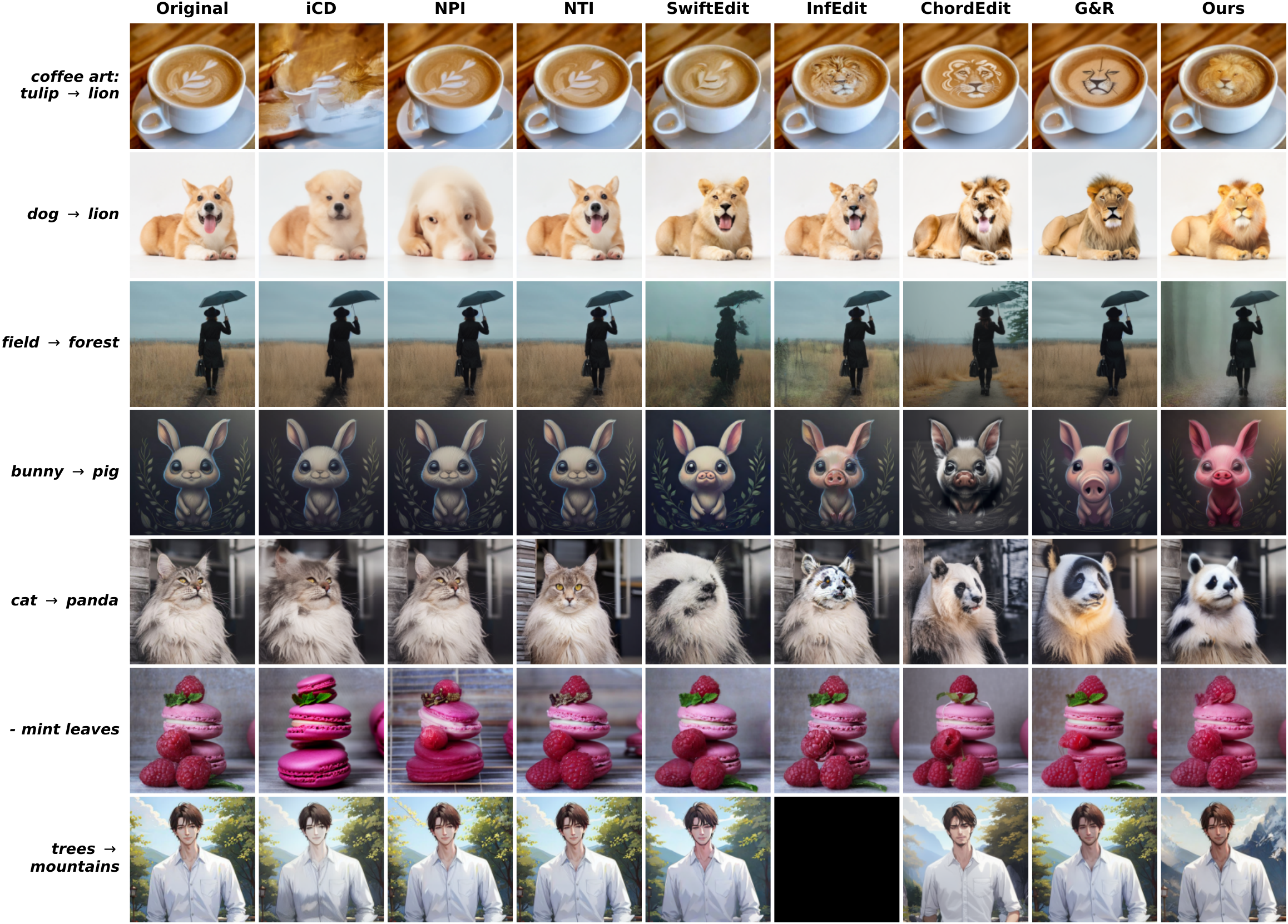}
    \caption{\textbf{Additional qualitative results on SD1.5 (LCM-DreamShaper)} against iCD,
NPI, NTI, SwiftEdit, InfEdit, ChordEdit, and G\&R (one edit per row).}
    \label{fig:editing_sd_appendix}
\end{figure}

We provide further qualitative results on each backbone family in
Figures~\ref{fig:editing_flow_appendix}--\ref{fig:editing_sd_appendix}: Z-Image Turbo against the flow-based
editors (Fig.~\ref{fig:editing_flow_appendix}), SDXL-Turbo against the SDXL editors
(Fig.~\ref{fig:editing_sdxl_appendix}), and SD1.5 against the SD editors (Fig.~\ref{fig:editing_sd_appendix}).
Across diverse edit types, including object replacement, material and style change,
attribute edits, and scene additions, our method applies the requested change while
preserving layout, identity, and background, whereas the baselines tend either to
under-edit or to drift from the source. Baselines are run with official code and default settings; for GNRi we additionally swept the edit CFG over [1,5] and report its best setting.
\subsection{Comparison with optimization-based few-step inversion}
\label{app:optinv}

PreciseInv~\cite{zuo2025precise} and TFinv~\cite{Wu_2026} refine the
inverted latent by per-image test-time optimization (${\sim}600$
optimization steps per image), making inference $6$--$16\times$ slower than our amortized inversion
for PreciseInv and ${\approx}275\times$ slower for TFinv (Table~\ref{tab:optinv}). We ran both with the
official code on the full 700-image PIE-Bench, selecting each
baseline's CFG by a calibration sweep; Table~\ref{tab:optinv} reports
the official protocol together with dCLIP and ImageReward.
PreciseInv attains slightly higher CLIP similarities, but our method
leads on directional alignment (dCLIP), ImageReward, and every
preservation metric on both backbones. This also answers whether per-image optimization upper-bounds our
approach: it does not, since the objective these methods optimize at
test time is reconstruction of the source, which constrains neither
alignment with the edit direction nor preservation once the prompt is
changed. Training through the full inversion--editing cycle optimizes
exactly those quantities, which is why the amortized inverter leads
on dCLIP, ImageReward, and all preservation metrics (trailing only on
raw CLIP similarity) while being up to two orders of magnitude
faster.

\begin{table}[t]
\centering\small
\setlength{\tabcolsep}{4pt}
\caption{Optimization-based few-step inversion vs.\ ours under the
official PIE-Bench protocol (dCLIP and Image Reward from our scorer).
Both baselines optimize per image at test time; ours is amortized.
Time is per-image inference on the same H100 GPUs as Table~\ref{tab:main_results}; the
baselines' per-image optimization early-stops, hence faster than the
nominal step count.}
\label{tab:optinv}
\begin{tabular}{lccccccccc}
\toprule
 & CLIP\textsubscript{wh}$\uparrow$ & CLIP\textsubscript{ed}$\uparrow$ & dCLIP$\uparrow$ & Struct$\downarrow$ & LPIPS$\downarrow$ & SSIM$\uparrow$ & PSNR$\uparrow$ & IR$\uparrow$ & Time$\downarrow$ (s) \\
\midrule
\multicolumn{10}{l}{\textit{SD-based}}\\
PreciseInv    & \textbf{26.65} & \textbf{22.95} & 0.129 & 33.0 & 0.132 & 0.765 & 21.9 & 0.520 & 3.4 \\
TFinv         & 24.53 & 21.48 & 0.072 & 45.0 & 0.133 & 0.744 & 19.8 & 0.036 & 165 \\
\textbf{Ours} & 26.13 & 22.47 & \textbf{0.155} & \textbf{22.8} & \textbf{0.071} & \textbf{0.843} & \textbf{23.6} & \textbf{0.573} & \textbf{0.6} \\
\midrule
\multicolumn{10}{l}{\textit{SDXL-based}}\\
PreciseInv    & \textbf{27.49} & \textbf{24.08} & 0.151 & 64.6 & 0.213 & 0.735 & 18.2 & 0.736 & 9.4 \\
\textbf{Ours} & 27.17 & 23.31 & \textbf{0.178} & \textbf{43.0} & \textbf{0.105} & \textbf{0.808} & \textbf{20.8} & \textbf{0.858} & \textbf{0.6} \\
\bottomrule
\end{tabular}
\end{table}
\subsection{Metric circularity and held-out re-scoring}
\label{app:heldout_clip}

Our training objective uses two pretrained networks: CLIP ViT-L/14
(directional loss, Eq.~\ref{eq:dclip_loss}) and VGG-16 (LPIPS reconstruction loss,
Eq.~\ref{eq:rec_loss}). All metrics of the official protocol are fixed by the
benchmark and applied identically to every method; in particular, its
LPIPS uses SqueezeNet, not the VGG-16 of our loss, Structure Distance
uses DINO ViT-B/8, and SSIM, PSNR, and MSE are model-free. Of the
scores we add to the protocol (dCLIP, Image Reward, runtime), only
dCLIP shares a model with training: its ViT-L/14 encoder. To rule out
encoder-specific bias we re-score every method in
Table~\ref{tab:main_results} with OpenCLIP ViT-bigG/14, an encoder
unused in training with a different pretraining corpus.
Table~\ref{tab:heldout} shows that within-block orderings are
identical and relative gaps persist under the held-out encoder; Image
Reward is an independently trained reward model, and the human study
(Appendix~\ref{app:human}) involves no model at all.

\subsection{Per-category comparison}
\label{app:percat}

Table~\ref{tab:percat} breaks the comparison down by the ten PIE-Bench
edit categories, reporting the within-block Avg Rank (computed as in
Table~\ref{tab:main_results}) of our method against the best
accelerated competitor. Ours ranks first in 24 of 30
backbone$\times$category cells (26 with ties; 21/30 when the full-step
SD editors are also included). The hardest category is color-attribute
editing on SD/SDXL: the strong recolor (dCLIP $0.26$--$0.28$) tends to
delocalize, a failure mode absent on the flow backbone (per-category
preserve-LPIPS $0.22$/$0.34$ vs.\ $0.15$).

\subsection{Masked-inference variant}
\label{app:masked}

The directional CLIP objective (Eq.~\ref{eq:dclip_loss}) is global, so
strong edits can spill onto semantically related content. We evaluate
an inference-time masked variant on the SD backbone: cross-attention
maps of the edit tokens define an edit-region mask, and the output
blends the edited image inside the mask with the source reconstruction
outside; no additional training is required. Table~\ref{tab:masked}
uses the same checkpoint and guidance scale as the ``Ours'' row of
Table~\ref{tab:main_results}, so masking is the only changed variable.
The mask trades edit strength for markedly better preservation:
Structure Distance improves from $22.8$ to $14.3$ and SSIM becomes the
best in the SD block, while dCLIP ($0.120$) still exceeds every SD
baseline in Table~\ref{tab:main_results}.

\begin{table}[t]
\centering\small
\setlength{\tabcolsep}{4pt}
\caption{Masked-inference variant on the SD backbone (same checkpoint
and $\omega{=}2.0$ as Table~\ref{tab:main_results}); official
PIE-Bench protocol, dCLIP and Image Reward from our scorer. dCLIP ($0.120$) still matches or exceeds every SD baseline
(the closest, ChordEdit, scores $0.119$)}
\label{tab:masked}
\begin{tabular}{lcccccccc}
\toprule
 & CLIP\textsubscript{wh}$\uparrow$ & CLIP\textsubscript{ed}$\uparrow$ & dCLIP$\uparrow$ & Struct$\downarrow$ & LPIPS$\downarrow$ & SSIM$\uparrow$ & PSNR$\uparrow$ & IR$\uparrow$ \\
\midrule
Ours          & \textbf{26.13} & \textbf{22.47} & \textbf{0.155} & 22.8 & 0.071 & 0.843 & 23.6 & \textbf{0.573} \\
Ours (masked) & 25.43 & 21.85 & 0.120 & \textbf{14.3} & \textbf{0.051} & \textbf{0.864} & \textbf{25.8} & 0.440 \\
\bottomrule
\end{tabular}
\end{table}
\begin{table}[t]
\centering\small
\setlength{\tabcolsep}{3.5pt}
\caption{Per-category Avg Rank vs.\ accelerated editors (lower is
better): ours against the best competitor in each cell. Bold marks the
cell winner; $=$ marks ties.}
\label{tab:percat}
\resizebox{\linewidth}{!}{%
\begin{tabular}{l cc cc cc}
\toprule
 & \multicolumn{2}{c}{SD} & \multicolumn{2}{c}{SDXL} & \multicolumn{2}{c}{Flow} \\
\cmidrule(lr){2-3}\cmidrule(lr){4-5}\cmidrule(lr){6-7}
Category & Ours & Best comp. & Ours & Best comp. & Ours & Best comp. \\
\midrule
Random        & \textbf{2.33} & 2.44 (InfEdit) & \textbf{1.89} & 2.11 (TurboEdit) & \textbf{1.33} & 2.67 (UniEdit) \\
Change object & \textbf{2.22} & 2.33 (InfEdit) & \textbf{1.67} & 2.22 (TurboEdit) & \textbf{2.33} & 2.67 (FlowEdit) \\
Add object    & \textbf{1.44} & 2.89 (InfEdit) & \textbf{1.56} & 2.33 (TurboEdit) & \textbf{1.56} & 2.67 (UniEdit) \\
Delete object & \textbf{2.00} & 2.67 (InfEdit) & \textbf{1.78} & 2.22 (GNRi)      & $=$2.78 & 2.78 (FlowEdit) \\
Content       & \textbf{1.44} & 2.67 (InfEdit) & \textbf{1.78} & 2.11 (TurboEdit) & \textbf{1.56} & 2.67 (UniEdit) \\
Pose          & 2.44 & \textbf{1.89} (InfEdit) & \textbf{1.67} & 2.22 (TurboEdit) & \textbf{2.11} & 2.89 (UniEdit) \\
Color         & $=$2.67 & 2.67 (InfEdit) & 2.11 & \textbf{2.00} (TurboEdit) & \textbf{1.56} & 2.67 (UniEdit) \\
Material      & \textbf{2.11} & 2.44 (InfEdit) & \textbf{1.67} & 2.22 (TurboEdit) & \textbf{1.33} & 2.22 (UniEdit) \\
Background    & 2.44 & \textbf{2.22} (InfEdit) & 2.22 & \textbf{1.89} (TurboEdit) & \textbf{1.78} & 2.67 (UniEdit) \\
Style         & \textbf{2.22} & 2.56 (InfEdit) & \textbf{1.33} & 2.33 (TurboEdit) & \textbf{1.56} & 1.78 (UniEdit) \\
\bottomrule
\end{tabular}%
}
\end{table}

\begin{table}[t]
\centering\small
\setlength{\tabcolsep}{5pt}
\caption{dCLIP under the training encoder (ViT-L/14) and a held-out
encoder (OpenCLIP ViT-bigG/14). Within-block ordering is identical.}
\label{tab:heldout}
\resizebox{\linewidth}{!}{%
\begin{tabular}{lcc @{\hskip 2em} lcc @{\hskip 2em} lcc}
\toprule
\multicolumn{3}{c}{\textit{SD-based}} & \multicolumn{3}{c}{\textit{SDXL-based}} & \multicolumn{3}{c}{\textit{Flow-based}} \\
Method & L/14 & bigG & Method & L/14 & bigG & Method & L/14 & bigG \\
\midrule
\textbf{Ours} & \textbf{0.155} & \textbf{0.183} & \textbf{Ours} & \textbf{0.178} & \textbf{0.231} & \textbf{Ours} & \textbf{0.154} & \textbf{0.193} \\
ChordEdit & 0.119 & 0.145 & TurboEdit & 0.145 & 0.187 & UniEdit-Flow & 0.123 & 0.156 \\
G\&R      & 0.106 & 0.123 & ReNoise   & 0.124 & 0.163 & FireFlow     & 0.117 & 0.151 \\
InfEdit   & 0.096 & 0.116 & GNRi      & 0.038 & 0.043 & FlowEdit     & 0.108 & 0.139 \\
NPI       & 0.085 & 0.107 &           &       &       & RF-Edit      & 0.103 & 0.128 \\
SwiftEdit & 0.081 & 0.096 &           &       &       &              &       &       \\
NTI       & 0.050 & 0.059 &           &       &       &              &       &       \\
iCD       & 0.047 & 0.055 &           &       &       &              &       &       \\
\bottomrule
\end{tabular}%
}
\end{table}


\begin{table}[t]
\centering\small
\setlength{\tabcolsep}{4pt}
\caption{Loss-weight sweeps on Z-Image-Turbo (step 600, guided
inference of Eq.~\ref{eq:guided_backward} at $\omega{=}2.5$; internal scorer as in Table~\ref{tab:ablation}). Each row is a separately
trained run; the $\lambda{=}0$ rows and the bold defaults coincide
with the corresponding rows of Table~\ref{tab:ablation}.}
\label{tab:loss_weight_sweeps}
\begin{tabular}{llcccccc}
\toprule
Sweep & $\lambda$ & CLIP$_{\text{wh}}\uparrow$ & dCLIP$\uparrow$ & LPIPS$\downarrow$ & SSIM$\uparrow$ & PSNR$\uparrow$ & IR$\uparrow$ \\
\midrule
\multirow{5}{*}{$\lambda_{\mathrm{LF}}$ ($\lambda_{\mathrm{KL}}{=}0.1$)}
 & 0             & 26.13 & 0.1423 & 0.2071 & 0.8171 & 21.49 & 0.6015 \\
 & 0.25          & 26.66 & 0.1641 & 0.2067 & 0.8062 & 20.71 & 0.8302 \\
 & 0.5           & 26.94 & 0.1677 & 0.2269 & 0.7922 & 20.21 & 0.8626 \\
 & \textbf{1}    & \textbf{26.44} & \textbf{0.1541} & \textbf{0.1959} & \textbf{0.8296} & \textbf{22.01} & \textbf{0.7383} \\
 & 2             & 27.01 & 0.1714 & 0.2636 & 0.7586 & 18.93 & 0.8501 \\
\midrule
\multirow{4}{*}{$\lambda_{\mathrm{KL}}$ ($\lambda_{\mathrm{LF}}{=}1$)}
 & 0             & 26.06 & 0.1350 & 0.1901 & 0.8306 & 21.81 & 0.6029 \\
 & 0.05          & 27.04 & 0.1638 & 0.2396 & 0.7804 & 18.73 & 0.8525 \\
 & \textbf{0.1}  & \textbf{26.44} & \textbf{0.1541} & \textbf{0.1959} & \textbf{0.8296} & \textbf{22.01} & \textbf{0.7383} \\
 & 0.5           & 16.24 & 0.0561 & 0.5860 & 0.3908 & 8.18 & $-$2.2740 \\
\bottomrule
\end{tabular}
\end{table}
\end{document}